\documentclass[]{adobe_asml}

\usepackage{blindtext}

\definecolor{adbered}{HTML}{f01404}

\usepackage{tcolorbox}
\usepackage{dblfloatfix}
\definecolor{cvprblue}{rgb}{0.21,0.49,0.74}
\usepackage{subcaption}
\usepackage{tikz}
\usetikzlibrary{patterns}

\usepackage{tcolorbox}
\usepackage{pdflscape}

\title{\textbf{
UniFusion: Vision-Language Model as Unified Encoder in Image Generation
}}
\runningtitle{Vision-Language Model as Unified Encoder in Image Generation}

\author{%
\parbox{\textwidth}{ \centering
Kevin (Yu-Teng) Li$^{*}$, Manuel Brack$^{*}$, Sudeep Katakol, Hareesh Ravi, Ajinkya Kale
}}

\affiliation{%
\parbox{\textwidth}{\centering\small
Adobe Applied Research
}}

\footnotetext[1]{Authors contributed equally. Kevin led the model design, training, and ablation experiments. Manuel led the evaluation, presentation, and writing of the paper.}
\usepackage{tikz}
\usetikzlibrary{patterns}

\abstract{

Although recent advances in visual generation have been remarkable, most existing architectures still depend on distinct encoders for images and text. This separation constrains diffusion models’ ability to perform cross-modal reasoning and knowledge transfer. Prior attempts to bridge this gap often use the last layer information from VLM, employ multiple visual encoders, or train large unified models jointly for text and image generation, which demands substantial computational resources and large-scale data, limiting its accessibility. 

To maximize the benefits of the joint multimodal reasoning and representation capacity of VLMs, we present \model, a diffusion-based generative model conditioned on a frozen large vision-language model (VLM) that serves as a unified multimodal encoder. At the core of \model~is the Layerwise Attention Pooling (LAP) mechanism that extracts both high level semantics and low level details from text and visual tokens of a frozen VLM to condition a diffusion generative model. We demonstrate that LAP outperforms other shallow fusion architectures on text-image alignment for generation and faithful transfer of visual information from VLM to the diffusion model which is key for editing. We propose VLM-Enabled Rewriting Injection with Flexibile Inference (\rewrite), which conditions a diffusion transformer (DiT) only on the text tokens generated by the VLM during in-model prompt rewriting. \rewrite~combines the alignment of the conditioning distribution with the VLM's reasoning capabilities for increased capabilities and flexibility at inference. With an 8B VLM and an 8B DiT, \model~surpasses Flux.1 [dev] and BAGEL on DPG-Bench with a smaller training set (<1 billion samples), while comparing favorably against Flux.1 Kontext [dev] and Qwen-Image-Edit in editing tasks without any post-training. In addition, finetuning on editing task not only improves text-image alignment for generation, indicative of cross-modality knowledge transfer, but also exhibits tremendous generalization capabilities. Our model when trained on single image editing, zero-shot generalizes to multiple image references further motivating the unified encoder design of \model.

}
\date{\today}
\checkdata[Correspondence]{\textcolor{adbered}{<yutengl, mbrack>@adobe.com}}
\checkdata[Project Page]{\url{https://thekevinli.github.io/unifusion/}}

\newcommand{\model}{\textsc{UniFusion}}
\newcommand{\rewrite}{\textsc{Verifi}}
\begin{document}
\maketitle

\section{Introduction}\label{sec:intro}

\begin{figure}
    \centering
    \includegraphics[width=.99\linewidth]{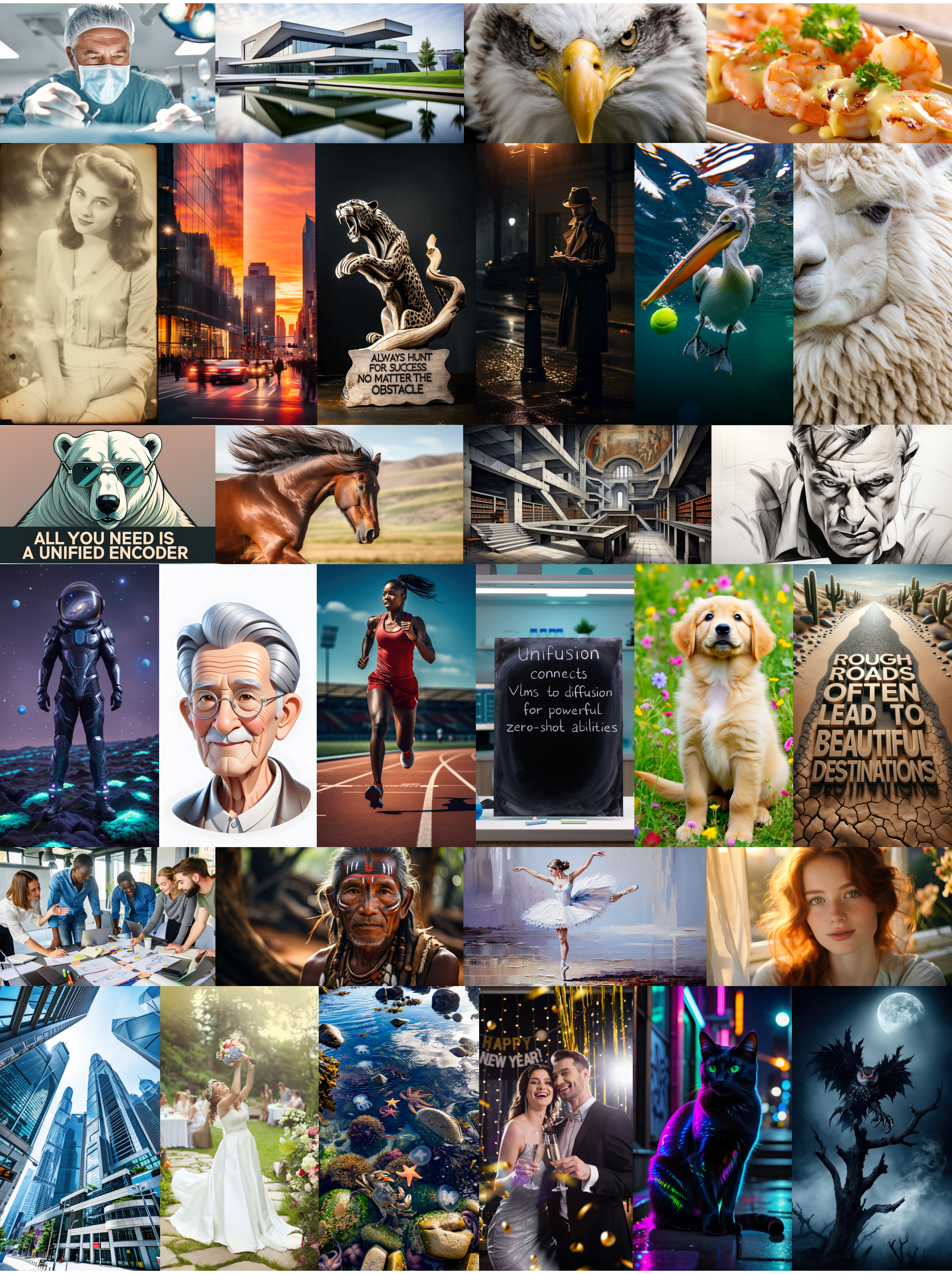}
    \caption{Diverse text-to-image generation with \model. (Zoom in for more details)}
    \label{fig:t2i_teaser}
\end{figure}

\begin{figure}
    \centering
    \includegraphics[width=.99\linewidth]{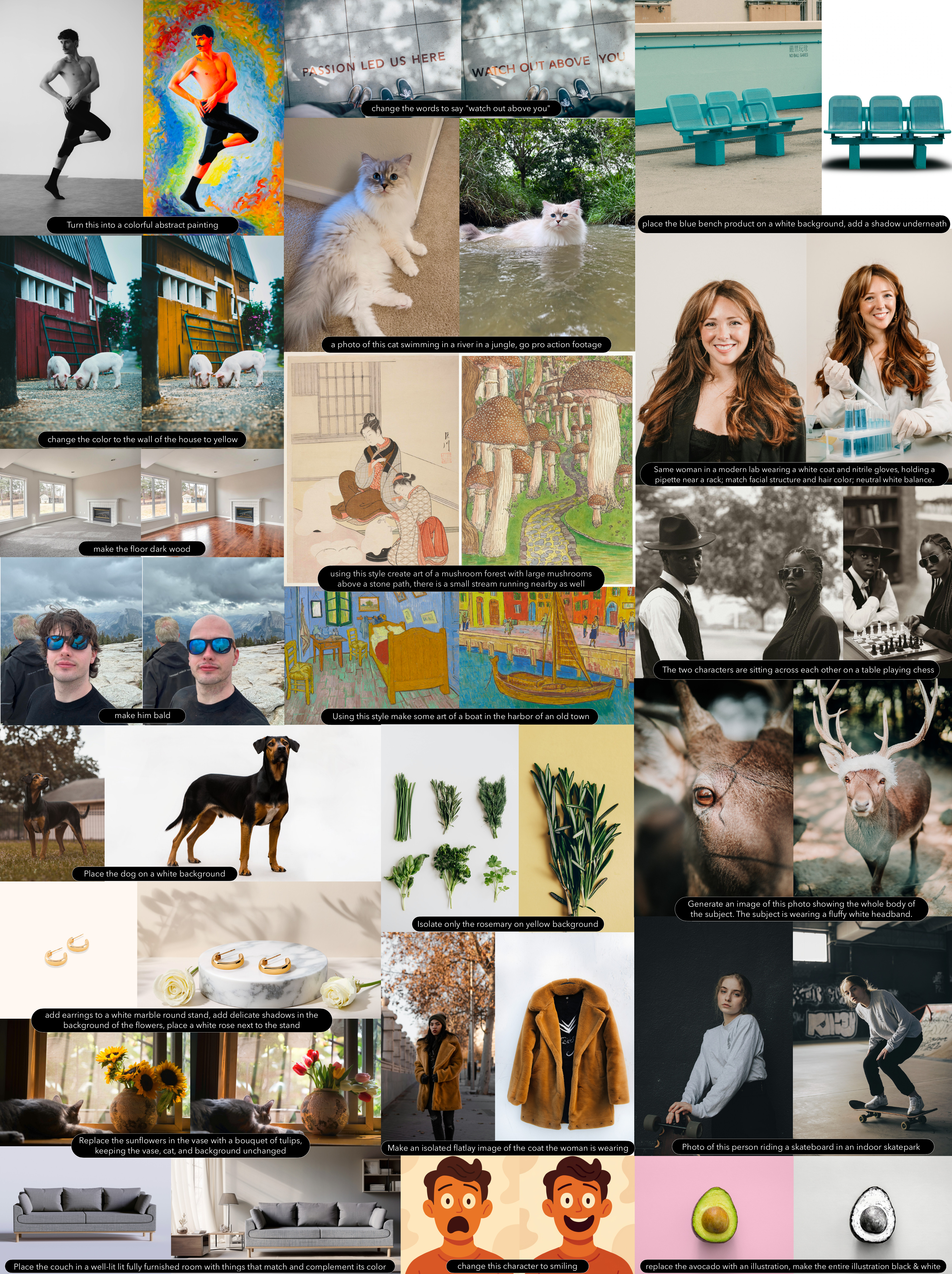}
    \caption{Diverse textual image editing and image reference workflows with \model. All images encoded by VLM features only, no VAE tokens involved. (Zoom in for more details)}
    \label{fig:ie_teaser}
\end{figure}

 \begin{figure}[t!]
     \centering
     
        \begin{subfigure}[t]{\textwidth}
            \centering
            \includegraphics[width=.85\linewidth]{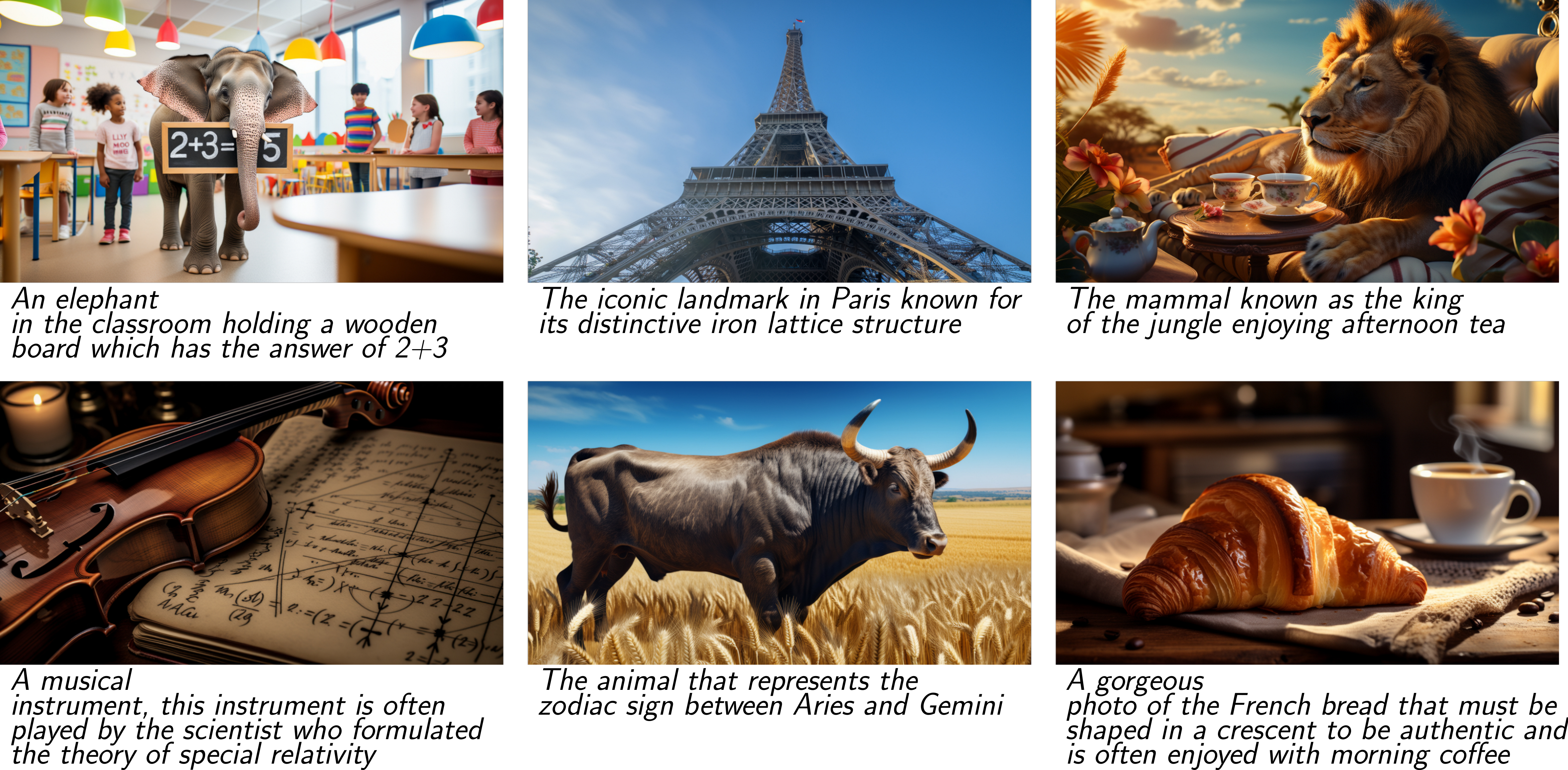}
            \caption{Zero-shot reasoning. Our \rewrite~paradigm allows \model~to leverage the world knowledge and reasoning of the VLM encoder.}
            \label{fig:teaser_t2ir}
        \end{subfigure}%
        \vspace{15px}
        \begin{subfigure}[t]{\textwidth}
        \centering
        \includegraphics[width=.98\linewidth]{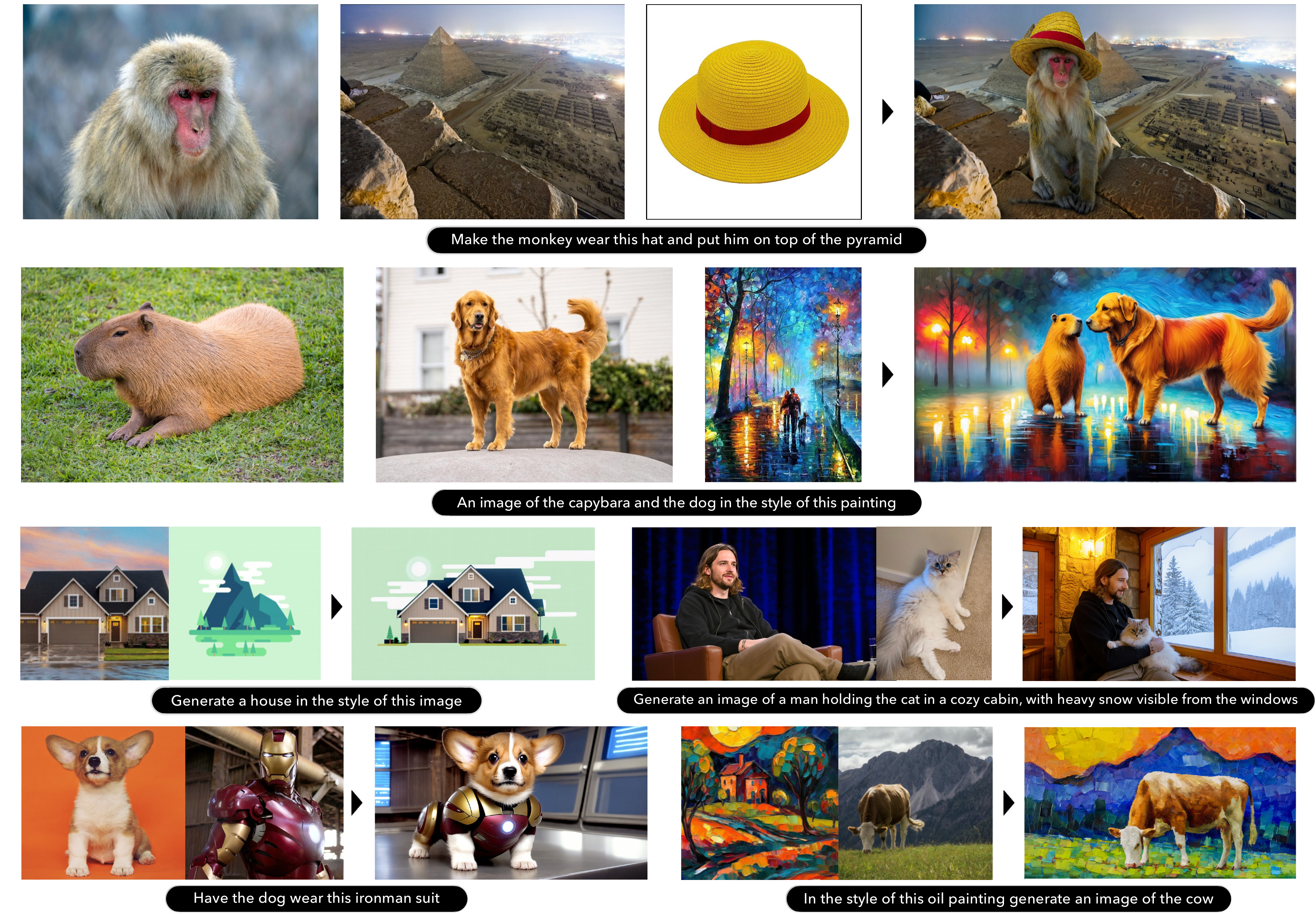}
        \caption{Zero-shot multi-reference generations for \model~model only trained on a single-reference samples.}
        \label{fig:zershot_multimage}
        \label{fig:teaser_imageref}
    \end{subfigure}
     \caption{Zero-shot capabilities by \model, which was not explicitly trained for. Our unified encoder setup enables the transfer of many capabilities from the VLM encoder to generative image applications.
     }
     \label{fig:teaser_zeroshot}

 \end{figure}

\begin{figure}
    \centering
    \includegraphics[width=.99\linewidth]{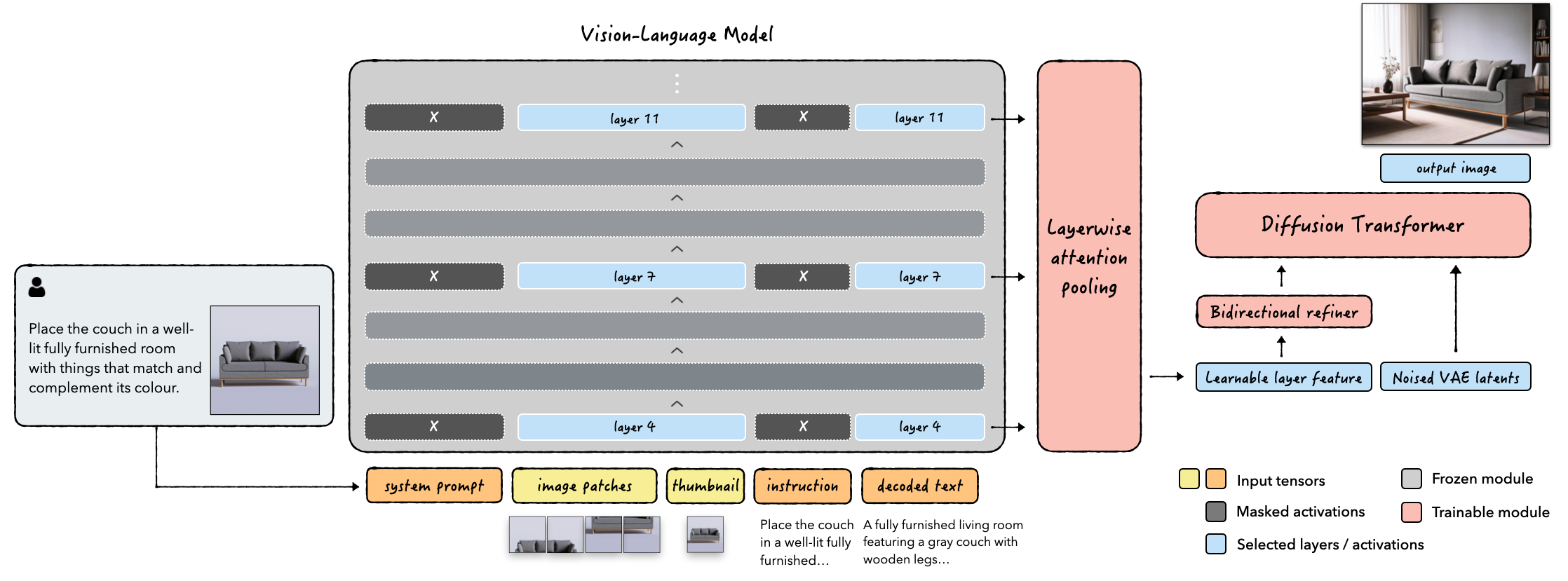}
    \caption{\model~architecture and inference paradigm. We extract multimodal representations from multiple layers of a frozen LLM and aggregate with a learnable layerwise attention pooling (LAP) module. A subsequent refiner counteracts the VLM's position bias due to causal attention. VLM-Enabled Rewriting Injection with Flexible Inference (\rewrite) rewrites the original input in-context. The rewritten tokens used for DiT conditioning leverage the VLM's reasoning capabilities to contextualize the target scene into a unified representation. }
    \label{fig:final_arch}
\end{figure}

The rapid advancement of generative image models has had a profound impact on creative tasks.
However, recent creative workflows demand models that go beyond text-to-image generation to support editing, reference-based composition, and iterative instruction-following.
While natively multimodal systems have emerged to handle such tasks \cite{yu2023cm3leon, zhou2024transfusion, chen2025janus, xie2024show, tian2025unigen, deng2025bagel}, they require joint training over both text and image modalities. This setting significantly increases computational and data requirements, potentially leading to adverse effects on image fidelity.
In this work, we focus on developing an image-generation model that achieves these capabilities without the complexity of joint multimodal training.

Current image-generation models typically condition on separate representation spaces for text and image inputs. Most commonly, T5 embeddings \cite{chung2024scaling} for text and variational auto-encoder (VAE) latents for images \cite{team2025seedream4, labs2025flux1kontext}.
However, these encoders operate at fundamentally different levels of abstraction: T5 captures high-level semantic meaning presented via text prompts while VAEs preserve low-level pixel-level detail from images. 
This mismatch is evident concretely in editing tasks, where models struggle to balance content preservation with instruction adherence, often producing either unnatural copy-paste artifacts or excessive modifications \cite{wang2025seededit3}.
We argue that separate encoding spaces force the DiT to expend capacity aligning heterogeneous features rather than synthesizing images, and that a unified semantic space can alleviate this burden.
VLMs naturally offer such a shared representation for both text and images, but prior works conditioning on VLM features report failure to preserve the fine-grained visual details required for high-fidelity editing \cite{bellagente2023multifusion, qwen-image}.

We propose \model~(Fig.~\ref{fig:final_arch}), a framework for building image-generation models with unified text and image encoding.
A frozen VLM serves as a unified encoder for both modalities, eliminating the need for separate conditioning spaces.
The framework comprises two key components: (1) Layerwise Attention Pooling (LAP), which aggregates information across multiple VLM layers to capture both fine-grained visual details and high-level semantic abstractions, and (2) VLM-Enabled Rewriting Injection with Flexible Inference (\rewrite), which only exposes the DiT a rewritten target prompt based on the original user input. \rewrite~reduces distribution shift between different input prompt formats and incorporates VLM's reasoning capabilities and world knowledge into the representations.

We demonstrate the effectiveness of the \model~framework by training a single model that achieves competitive performance on both text-to-image generation and editing compared to strong contemporaries, without requiring any supervised fine-tuning or reinforcement learning.
Notably, the resulting model exhibits remarkable zero-shot generalization capabilities: it handles multi-reference image inputs despite being trained only on single-reference editing data, and can perform image-to-image variations when exclusively trained on text-conditional generation.
We further observe cross-task positive transfer, where training on editing tasks improves the model's text-to-image prompt adherence and aesthetic quality.


In Fig.~\ref{fig:t2i_teaser},  \ref{fig:ie_teaser}, and \ref{fig:teaser_t2ir}, we showcase exemplary use cases of \model~and the benefits of tight conditioning on a unified encoder. One single model enables 1) high-fidelity text-image-generation with strong prompt following for complex instructions, 2) reliably usage of reference images for content and style, 3) text-driven image editing, 4) strong (visual) reasoning for complicated tasks, 5) usage of multiple image inputs and references for complex use cases, 6) generalization to unseen tasks, such as multi-reference images and cross-aspect ratio object consistency.

Our contributions can be summarized as follows:
\begin{itemize}
    \item We propose \model, a framework for image generation with unified text and image encoding via a frozen VLM, comprising two key components: Layerwise Attention Pooling (LAP) for multi-layer feature aggregation and VLM-Enabled Rewriting Injection with Flexible Inference (\rewrite) for distribution alignment.
    \item We conduct extensive ablations across prominent conditioning strategies, demonstrating that LAP outperforms conventional last-layer extraction and alternative fusion schemes while maintaining architectural flexibility.
    \item We train and validate a single model that achieves competitive performance on both text-to-image generation and editing tasks using only VLM input features, eliminating the need for separate VAE-based image reference conditioning.
    \item We demonstrate zero-shot generalization capabilities, including multi-reference composition despite training only on single-reference data, and image-to-image variation despite training exclusively on text-conditional generation.
\end{itemize}

The rest of this work is structured as follows. In Sec.~\ref{sec:background}, we explore potential architectures for VLM conditioned image generation. We consider existing methods and propose novel approaches for extracting representations from multiple VLM layers. 
In a direct comparison of these approaches, LAP outperforms all other methods. 
However, switching to a unified VLM encoder requires additional considerations beyond current conditioning paradigms. Consequently, Sec.~\ref{sec:ablations} goes into more detail on further design choices. We apply all of our insights to a final \model~model which we introduce and evaluate in Sec.~\ref{sec:unifusion}.
Given the strong zero-shot capabilities we observed for \model, we dedicate Sec.~\ref{sec:emergent} to investigating these in detail before concluding.

\section{Architecture Selection}\label{sec:background}\label{sec:architecture}

In this section, we first formally introduce potential paradigms for VLM-conditioned image generation. We perform direct comparisons in which LAP~outperforms all other methods. Then, we evaluate the preservation of fine-grained image details through LAP and compare two different paradigms for feature injection.
\subsection{VLM conditioning candidates}
\begin{figure}[t]
\begin{minipage}{.65\textwidth}
        
     \centering
       
        \begin{subfigure}[t]{0.48\textwidth}
        \centering
        \includegraphics[trim={1cm 16cm 31cm 1.75cm},clip,width=.98\linewidth]{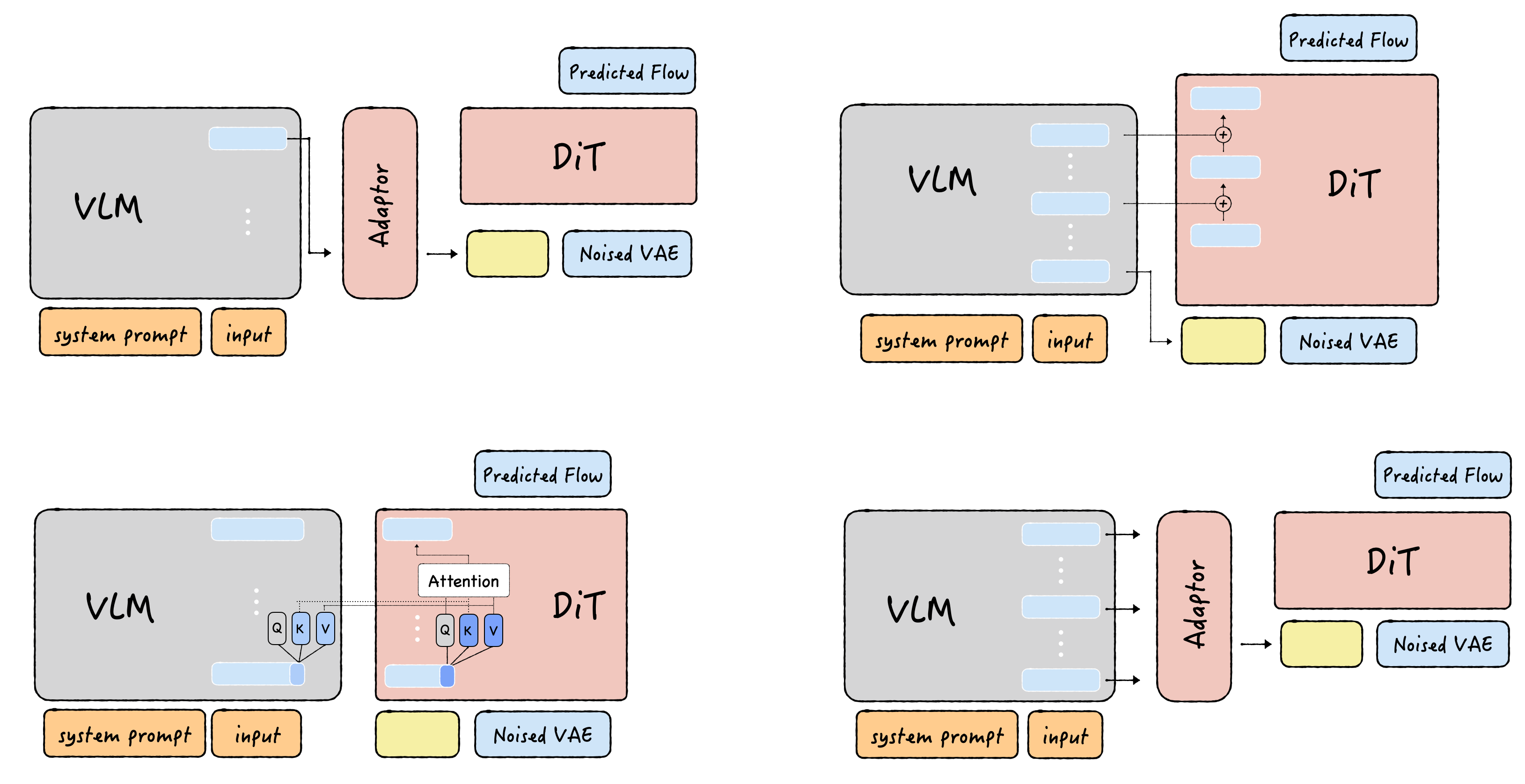}
        \caption{Last-Layer Hidden State}
        \label{fig:architecture_ll}
    \end{subfigure}%
    \hfill
    \begin{subfigure}[t]{0.48\textwidth}
        \centering
                \includegraphics[trim={1cm 1cm 33cm 15cm},clip,width=.98\linewidth]{figures/architecture_v2.png}
        \caption{Layerwise Key-Value Fusion}
        \label{fig:architecture_kv}
    \end{subfigure}
        \begin{subfigure}[t]{0.48\textwidth}
        \centering
        \includegraphics[trim={31cm 15cm 2cm 0.5cm},clip,width=.98\linewidth]{figures/architecture_v2.png}

        \caption{Hidden State Injection}
        \label{fig:architecture_hsi}
    \end{subfigure}%
    \hfill
    \begin{subfigure}[t]{0.48\textwidth}
        \centering
        \includegraphics[trim={31.5cm 0.75cm 0.5cm 16cm},clip,width=.98\linewidth]{figures/architecture_v2.png}
        \caption{Layerwise Attention Pooling (Ours)}
        \label{fig:architecture_lap}
    \end{subfigure}
        \caption{Overview of considered architectures for unified VLM conditioning. Blue blocks within the VLM and DiT module denote selected layers, red denotes trainable modules, and gray denotes frozen modules.}
        \label{fig:architecture}
            
\end{minipage}
\hfill
\begin{minipage}{.32\textwidth}
        \centering
    \includegraphics[width=0.95\linewidth]{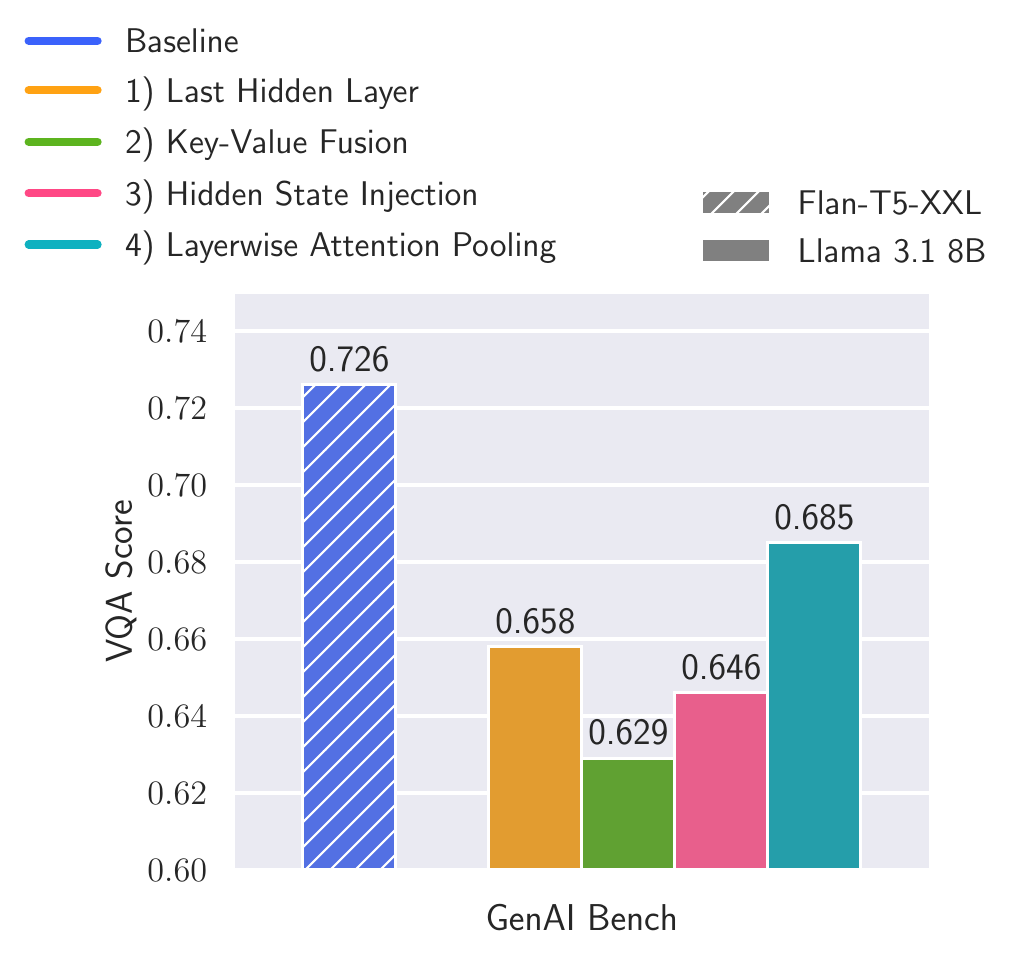}
    \caption{Comparison of different unified encoder candidates after 200k training steps on text-to-image performance (measured by VQA Score~\cite{lin2024evaluating} on GenAI-Bench~\cite{li2024genaibench}). LAP stands out as the best fusion strategy, whereas Key-Value Fusion consistently exhibits the lowest performance. Neither naive drop-in replacement of Llama-3.1 surpasses the T5 baseline. }
    \label{fig:text_to_image_ablations}
\end{minipage}
 \end{figure}

We consider four different architectural paradigms for a VLM-conditioned unified encoder as depicted in Fig.~\ref{fig:architecture}. For all approaches, we extract features from a frozen VLM that are used to condition a generative DiT.

\paragraph{Notation.}
We use the following notations to describe different methods. Consider the frozen VLM $E$ and trainable DiT $D$ with $N_E$ and $N_D$ layers, respectively. Here, $n=0$ corresponds to the input layer and $n=N$ to the last hidden state of the respective transformer.  
At any encoder layer $l^E_n$, we consider hidden states $x_n$ in the shape of \texttt{(bs, sl, h$_E$)}, denoting batchsize, sequence length, and the VLM's hidden dimension, respectively. Note that the token sequence will consist of system and user prompts and contains multimodal tokens for text and images. For simplicity, we abstract any transformer block to operation $Attn = \text{softmax}(Q K^T)V$, since details on scaling, multiple heads, and normalization are not affected by the considered methods.

\paragraph{Last-Layer Hidden State Encoding.} An intuitive approach is to extract representations from the last hidden layer of the VLM as a drop-in replacement for text conditioning in existing architectures (Fig.~\ref{fig:architecture_ll}). \citet{bellagente2023multifusion} proposed an early application of this method. More recently, multiple papers have similarly used the last hidden layer of a strong auto-regressive model \cite{qwen-image, xie2025sana, chen2025multimodal}. The most important design choice in this setup is the post-processing or pooling of the extracted representation. For example, \citet{bellagente2023multifusion} reported that they needed additional fine-tuning of the VLM to produce useful embeddings, while \citet{xie2025sana} only added an RMSNorm layer \cite{zhang2019rmsnorm}. Other variants of this approach have been proposed that use the penultimate layer instead of the last one \cite{qin2025lumina}. 

More formally, we extract conditioning $c$ as $c=x_N$ as the hidden state of the encoder layer $l^E_N$. An optional adapter $A(c) = c^\prime$ might project \texttt{h}$_D$ into the DiTs target dimension or implement additional normalization. $c^\prime$ is then concatenated with the noised VAE tokens $s_\text{vae}$, $c^\prime \oplus s_\text{vae}$. Subject to further embedding, this concatenated sequence is the input to the first DiT layer $l^D_0$. Consequently, the implementation of the DiT layers $l^D$ remains unaffected.

\paragraph{Layerwise Key-Value Fusion.}
One of the first proposed methods utilizing information from multiple layers is layer-wise Key-Value Fusion (Fig.~\ref{fig:architecture_kv}). \citet{liu2024playgroundv3improvingtexttoimage} proposed to match the number of layers and hidden dimension of the image generator to the encoder model. In each attention layer, we then concatenate the Keys and Values of the DiT with the respective Keys and Values of the encoder. 

Key value fusion requires that $N_E = N_D$ and $\texttt{h}_E = \texttt{h}_D$\footnote{Additionally, the number of number of attention heads, and attention heads dimensions need to match as well.}. For each layer $l_n, n \in N$ we extract $K_D$ and $V_D$ from the $Attn$ operation in the VLM. The $Attn$ operation in the respective DiT layers is adjusted such that $Attn = \text{softmax}(Q_D (K_D^T \oplus K_E^T))(V_D \oplus V_E)$ with concatenation $\oplus$ on the sequence dimension. Consequently, we still concatenate the encoder sequence $x_n$ with the sequence of noisy VAE tokens $s_\text{vae}$, but do so on every layer $n \in N$ and on the Attention Keys and Values instead of the residual stream between transformer blocks.

\paragraph{Hidden State Injection (HSI).}
We also consider an improvement over the previous approach that eliminates the need for Key-Value matching. Instead, we inject the representation from corresponding layers directly in the DiT through numerical addition of the residual stream after each block (Fig.~\ref{fig:architecture_hsi}). 

Again, we require that $N_E = N_D$ and $\texttt{hd}_E = \texttt{hd}_D$\footnote{However, the number of number of attention heads, and attention heads dimensions between Encoder and Decoder can differ.}. For each layer $l_n, n \in N$, we extract the hidden state $x^E_n$, which we add to the corresponding hidden state of the DiT $x^E_n$, such that ${x^E_n}^\prime = {x^E_n} + x^D_n$.

\paragraph{Layerwise Attention Pooling (LAP).}
We propose to aggregate information from intermediate layers using a learnable pooling module (Fig.~\ref{fig:architecture_lap}). LAP consists of 2 self-attention blocks that attend to the same token across layers, followed by a fully connected (FC) layer pooling the representations into one feature (See App.~Fig.~\ref{fig:unifusion_architecture}). 
This LAP setup can be flexibly integrated into the DiT architecture in various ways (Sec.~\ref{subsec:lap_injection}). 

For each layer $l^E_n, n \in N_E$  we extract the hidden state $x^E_n$. We then stack this tensor of shape (\texttt{bs, sl, n, h$_E$}) as $X^E$ (\texttt{bs*sl, n, h$_E$}). The LAP module consists of two standard transformer blocks with full self-attention on layers $n$: $X^{E\prime} = Attn(Attn(X^E)$. We then unstack $X^{E\prime}$ into its original shape and input it to a FC layer, such that $c^\prime = FC(X^{E\prime})$ of shape \texttt{bs, sl, h$_E$}. We can inject $c^\prime$ into the DiT $D$ as described for Last-Layer Injection. We also consider learning a dedicated LAP for each DiT layer $l^D_n, n \in N$ with each $c_n^\prime$ being injected via Hidden State Injection.

\paragraph{Benefits and Shortcomings of each Approach.}
The main limitation of Last-Layer Hidden State encodings is the restriction to representations from only one layer. Multiple prior works have established that transformer layers at different depths encode varying levels of information \cite{lindsey2025biology, durrani2020emnlp, durrani2023discovering, hennigen2020intrinsic}. Crucially, we argue that intermediate layers also carry different levels of \textit{semantic} abstraction and fine-grained details that are necessary for a unified encoder. Prior works have reported that the last-layer hidden state is insufficient in capturing detailed image contents and only provides a semantic abstraction \cite{bellagente2023multifusion, qwen-image}. 

While Layerwise Key-Value Fusion and Hidden State Injection extract features from multiple layers, they force tight coupling between the encoder VLM and the generative model, losing flexibility in the architectural design of the DiT. Since the number of layers and hidden dimensions is tied to the VLM, scaling the parameters of the generative model becomes challenging.

Conversely, LAP offers great flexibility in DiT architecture design while aggregating representations of different semantic granularity from the VLM.
\subsection{Experimental Evaluation}

\paragraph{Experimental Setup}
All direct comparisons in this section are conducted with the following setup to ensure fair comparisons. 
We utilize a standard latent DiT architecture with full self-attention and 2x2 patchification. The DiT has 32 layers with 32 attention heads and a hidden dimension of 4096, resulting in a 5 Billion parameter model. 

In line with previous work \cite{liu2024playgroundv3improvingtexttoimage, ma2024exploring}, we use frozen Llama3.1-8B \cite{grattafiori2024llama3} for text-to-image tasks.
Subsequently, we apply our findings to multimodal tasks using InternVL2.5-8B \cite{chen2025expanding}. We also train a model conditioned on Flan T5 XXL \cite{chung2024scaling} to serve as a baseline for text-to-image generation. 

For all experiments, we use the InstaFlow training objective \cite{liu2023instaflow} and AdamW optimizer \cite{Loshchilov2019decoupled}. We train on a global batch size of $1024$ for 200k steps, unless stated otherwise. While initially considered longer-running ablations, we observed that performance gaps at 200k steps serve as a reliable indicator of final model performance after extended training. We train the Llama and T5 checkpoints on text only, whereas the InternVL version sees a mix of 85\%/5\%/10\% text/image/text-image batches, respectively. For these ablations, we use paired image-caption data as multimodal training samples. In all our settings, the respective encoding model remains frozen. 

For text-to-image generation, we track the VQA score on GenAI bench \cite{li2024genaibench}. Through careful human comparison of pairwise VQA scores, we established the following evaluation setting as statistically meaningful. We observed VQA scores to correlate well with preferences in under-trained regimes below 80\%. In these settings, a gain of over 1 percentage points equates to noticeable performance improvement.

The image-to-image tasks mainly serve as a proxy for image encoding abilities; thus, we consider standard image difference metrics such as LPIPS \cite{zhang2018perceptual}, along with the more recent DreamSim metric \cite{fu2023dreamsim}.

\paragraph{Text-to-Image Prompt following}\label{subsec:encoder_knowledge}
In Fig.~\ref{fig:text_to_image_ablations}, we provide GenAI Bench performance of all architecture candidates. Overall, LAP stands out as the best option in terms of prompt adherence. The Llama 3.1 LAP version outperforms Last Hidden layer and HSI approaches. In comparison, Key-Value fusion performs significantly worse than all other methods. 
We argue that the prior success of this architecture \cite{liu2024playgroundv3improvingtexttoimage} can be mainly attributed to high-quality training captions, rather than the conditioning methodology. We hypothesize two reasons leading to the shortcomings of Key-Value Fusion. First, naive Key-Value concatenation without dedicated projections for each target layer is likely to lead to feature misalignment between the encoder and DiT. Second, the approach shares similarities with text-latent cross-attention in U-Nets, which we also found to be suboptimal compared to full self-attention text-conditioning.

While LAP emerged as the best candidate, no Llama-3 conditioned checkpoint reaches the performance of the T5 baseline. These results align with independent observations \cite{ma2024exploring} and can be attributed to multiple factors. 
For one, we observed T5 conditioned models to converge faster than Llama ones. After 400k training steps the gap in VQA score between the T5 and Llama-3 LAP model closes to $0.007$ percentage points (from $0.041$ points at 200k). Despite its prominent usage in prior works \cite{liu2024playgroundv3improvingtexttoimage, ma2024exploring}, we also found Llama to be a suboptimal encoder candidate. When using InternVL in the same setting, for example (Sec.~\ref{subsec:causual_attention}, we observed a significantly smaller gap to the T5 baseline.

Additionally, further manual inspection reveals that while the Llama conditioned model does better on prompt understanding in many examples, it fails to produce a key subject in the prompt for some samples. T5, in contrast, performs much more consistently.
The positional bias introduced by the causal attention masking of auto-regressive models is a major factor in this consistency gap \cite{ma2024exploring}.
Leveraging strong auto-regressive, decoder-only (multimodal) LLMs for conditioning in generative image tasks is thus not strictly plug-and-play but requires some additional adjustments over current paradigms (see Sec.~\ref{subsec:causual_attention}). 

Based on our Llama-3.1 ablations, we decided to move forward with LAP as the architecture for \model. It outperforms HSI and last hidden layer approaches, and provides higher flexibility than HSI with no inherent requirements on layer count or hidden dimension. We explore how to best utilize our LAP setting in Sec.~\ref{sec:ablations}, which ends up clearly outperforming T5 baselines on text-to-image generation and simultaneously supports further use cases.

\begin{figure}[t]
     \centering
        \begin{subfigure}[t]{0.6\textwidth}
        \centering
        \includegraphics[height=12.5em]{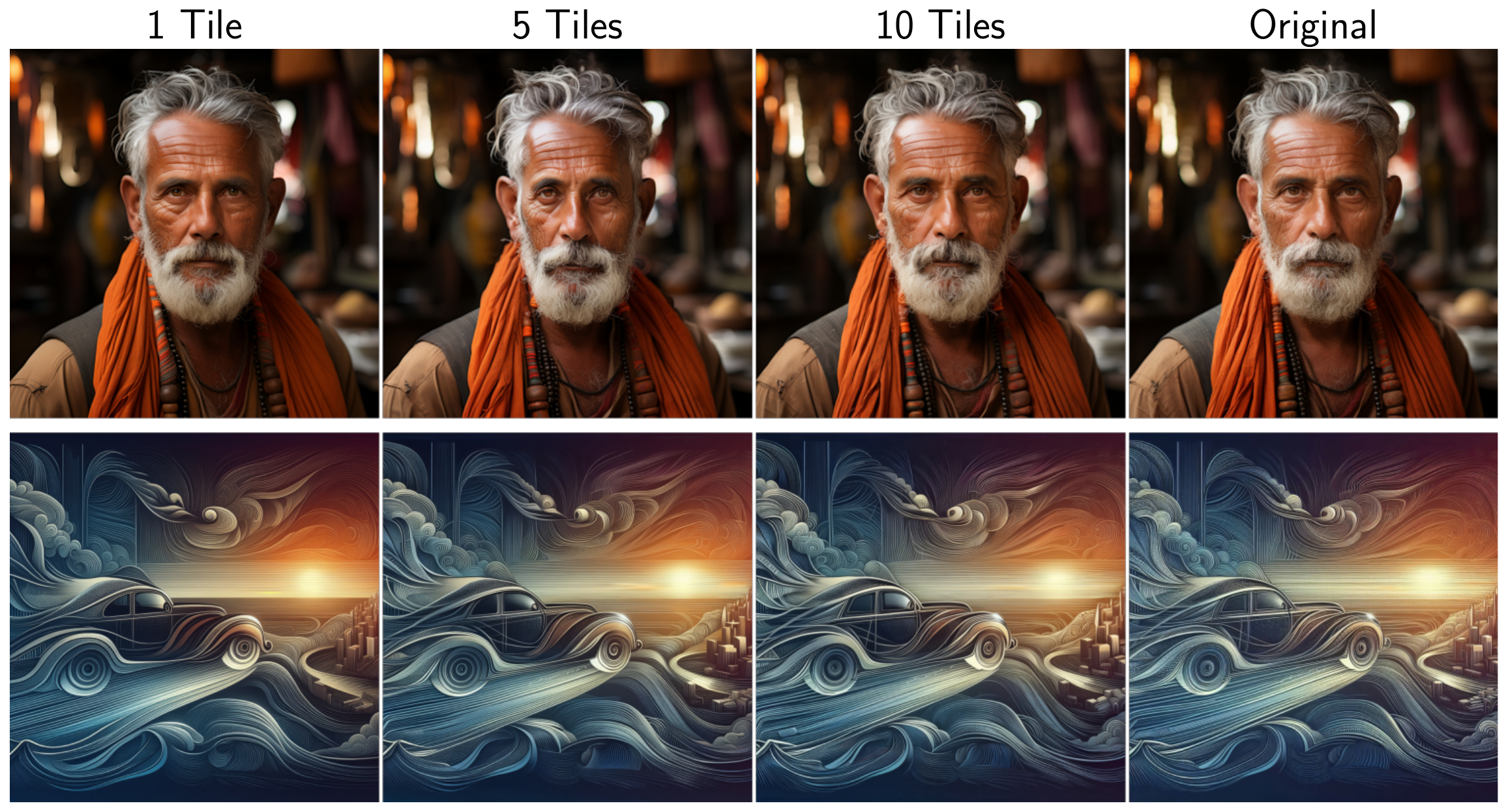}
        \caption{Qualitative Examples}
        \label{fig:image_rec_qual}
    \end{subfigure}%
    \hfill
    \begin{subfigure}[t]{0.38\textwidth}
        \centering
        \includegraphics[height=11.5em]{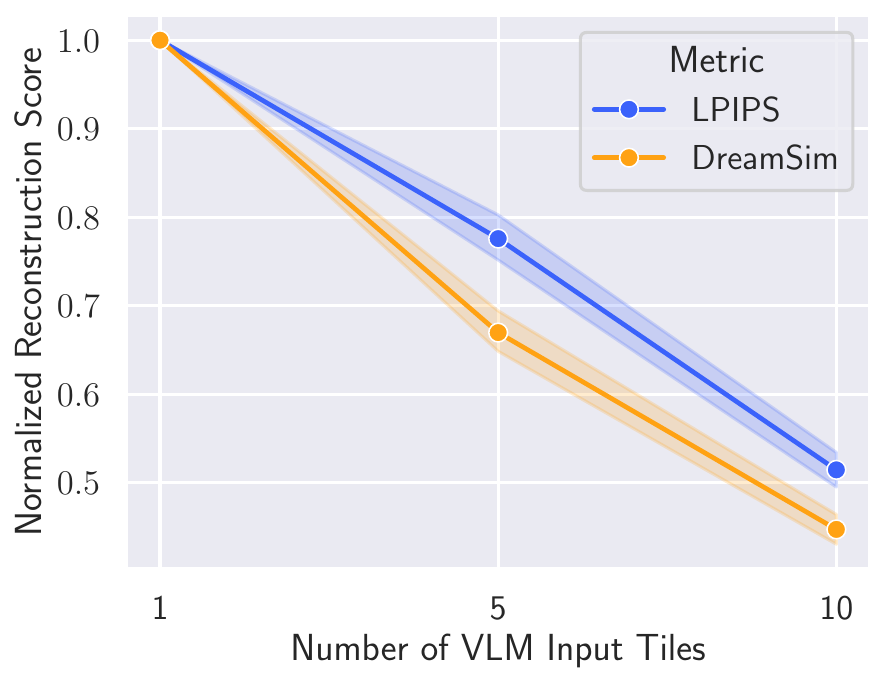}
        \caption{Quantitative Reconstruction Quality}
        \label{fig:image_rec_metric}
    \end{subfigure}
        \caption{Image reconstruction quality when the input image is patchified into 1, 5, and 10 tiles before being fed to the VLM. LAP extracted VLM features are capable of preserving input image details without additional feature injection.  Small, fine-grained images require more input image tiles for the VLM to be captured accurately. }
        \label{fig:image_rec}
 \end{figure}

\begin{tcolorbox}[leftrule=1.5mm,top=0.8mm,bottom=0.5mm,title=\model~Feature Extraction]
\begin{itemize}
    \item We extract features from multiple layers of the encoder model 
    \item We aggregate these activations using a Layerwise Attention Pooling (LAP) module consisting of two transformer blocks and a fully connected layer
\end{itemize}
\end{tcolorbox}

\paragraph{Image Information Preservation}\label{subsec:image_to_image}
With the benefits of LAP over other architectures established on text-to-image tasks, we shift our focus to image inputs. A unified encoder should be able to preserve fine-grained visual details to obtain precise edits, but previous work reported that VLM-based features specifically fall short of that hurdle \cite{bellagente2023multifusion, qwen-image}. 


In addition to the importance of utilizing features over multiple layers of the VLM, the representation capacity of the extracted features also plays an important role,
The number of image tokens at a given hidden dimension is often significantly lower than that of comparable VAEs, for example. Naturally, in such a setting, adding VAE-encoded image input tokens improves the preservation of fine-grained details. 

We compared different numbers of image tiles used in the image encoding of the VLM. As shown in Fig.~\ref{fig:image_rec}, the preservation of small features does indeed scale with the number of VLM tiles. At 10 tiles, any reconstruction errors become largely imperceptible. Even fine-grained structures, such as hairs or complex patterns, are preserved well. 
Thus, we conclude 
that VLM features are sufficient for image encoding. However, we need to accommodate a high number of tiles or image tokens and utilize features from earlier layers.

\begin{tcolorbox}[leftrule=1.5mm,top=0.8mm,bottom=0.5mm,title=\model~Image Input Encoding]
\begin{itemize}
    \item \model~only uses extracted VLM features to encode input and reference images 
    \item Increasing the number of image tiles in VLM image input encoding is crucial for preservation of fine-grained image details 
    \item Thus, \model~eliminates the need to add VAE encoded image tokens to the DiT input
\end{itemize}
\end{tcolorbox}

\paragraph{Representation Injection}\label{subsec:lap_injection}
When aggregating features with LAP, we are presented with different options on how and where to inject representations into the DiT. 

The two main options are: 1) to learn a dedicated LAP module for each DiT layer $l^D_n, n \in N$ for which we aim to inject features, and 2) Only extract a single pooled representation $c^\prime$, which we concatenate with noised VAE tokens $s_\text{vae}$ as input to the DiT $c^\prime \oplus s_\text{vae}$, similar to current conditioning approaches (See Sec.~\ref{sec:architecture}). 

We evaluate this design choice by comparing two models using LLama-3.1-8B \cite{grattafiori2024llama3} as an encoder. Here, the first model learns a dedicated LAP for each target injection layer of DiT, whereas the second uses a single representation injected in the DiT's input sequence. To control for total capacity across LAP modules, we scale the single LAP in the second setting to have a similar parameter count as all LAP modules in the first setting. 

In this direct comparison, the single, pooled LAP representation setting strongly outperforms its counterpart where LAP features are injected into DiT layers (App.~\ref{app:injection}). These results suggest that injecting conditioning into later layers of the DiT may be counterproductive, as we show in Sec.~\ref{subsec:layer_selection}. 

\begin{tcolorbox}[leftrule=1.5mm,top=0.8mm,bottom=0.5mm,title=\model~Feature Injection]
\begin{itemize}
    \item We use a single LAP module, converting all layer activations into one feature 
    \item These tokens are input as standard conditioning by pre-pending to the noisy VAE tokens in the DiT input sequence
\end{itemize}
\end{tcolorbox}

\section{\model~Design}\label{sec:ablations}

We have established Layerwise Attention Pooling (LAP) as the most promising conditioning strategy for a unified encoder in Sec.~\ref{sec:architecture}. In this section, we go into more detailed design choices of \model~and LAP.


\begin{figure}[b!]
    \centering
    \includegraphics[width=0.75\linewidth]{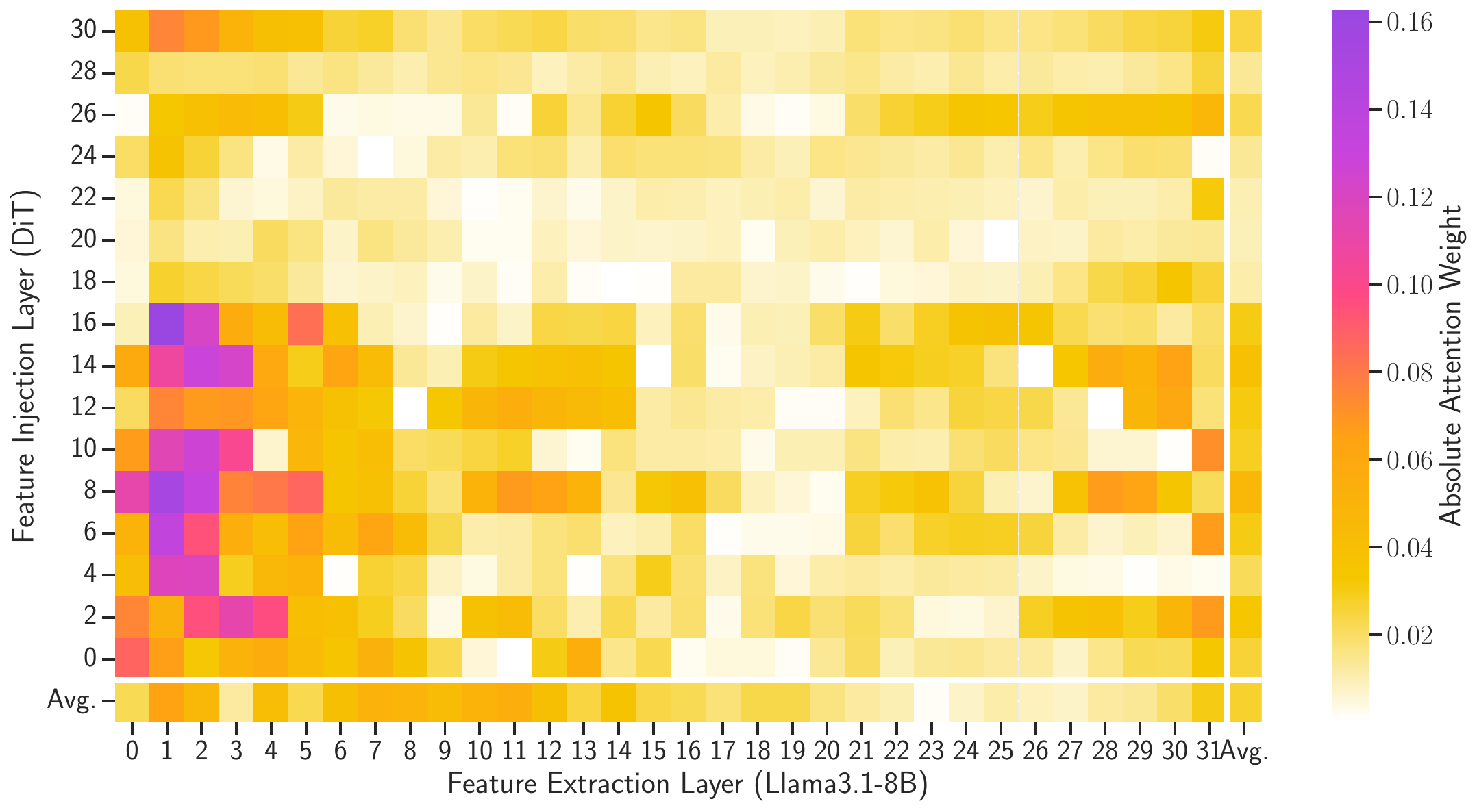}
    \caption{Weight visualization of LAP modules' pooling layers in Representation Injection setup (Sec~\ref{subsec:lap_injection}). Each value denotes the magnitude of weights assigned to each VLM layer at a given LAP module's pooling layer (smaller y-coordinate denotes layers closer to DiT input). On average, early VLM layers contribute more than later ones, while layer injection at later DiT blocks has lower weights. 
    }
    \label{fig:pooling_weights}
\end{figure}

\subsection{Layer Selection}\label{subsec:layer_selection}
While the previous results have shown clear benefits of aggregating representations from multiple layers, not all layers will be equally relevant, and information captured across different layers may be redundant. 
Thus, utilizing all layers may cause high memory overhead and potentially incentivize DiT to overfit on a small subset of layers instead of the full capacity of a VLM. 

We begin our analysis by visualizing the weights of the learnable pooling layer within LAP modules as shown in Fig.~\ref{fig:pooling_weights}.  We see that not all VLM layers contribute equally to the final representation. The model shows a clear tendency to allocate higher weights to early-to-middle VLM layers when given the freedom to do so. 
This observation aligns with the intuition that semantic information useful for downstream finetuning lives in the earlier part of a VLM.

\begin{figure}[t]
    \centering

        \centering
        \includegraphics[width=.98\linewidth]{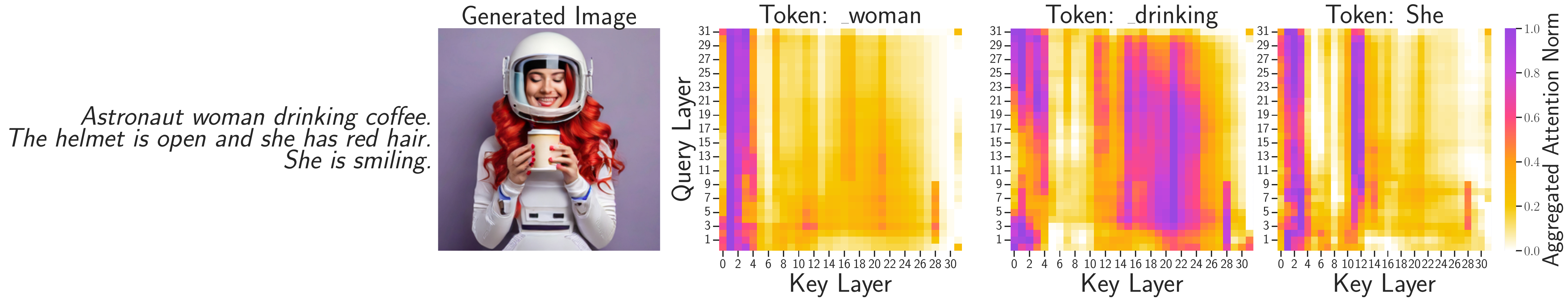}
        \caption{Qualitative example of local clusters in LAP Key-Value activations norm. On many tokens, the model utilizes implicit clusters of adjacent layers. Values are averaged over tokens if a word has more than 1 token.
        }
        \label{fig:local_clusters}
    \end{figure}

\begin{figure}[b!]
    \centering
    \begin{minipage}{.38\textwidth}
        \centering
\includegraphics[width=.95\linewidth]{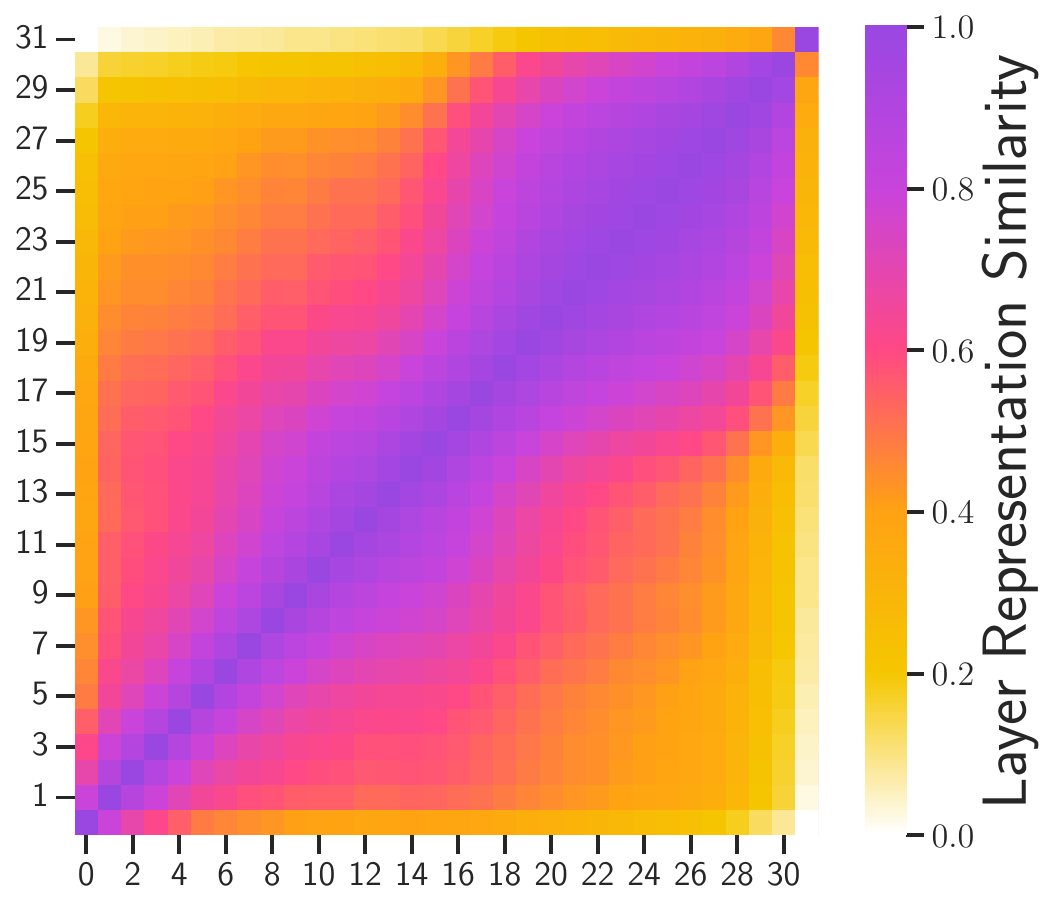}
        \caption{Adjacent layers in the VLM produce highly similar representations. Consequently, extracting features across each layer in the residual stream gives redundant information.}
        \label{fig:layer_similarity}
    \end{minipage}%
    \hfill
    \begin{minipage}{0.58\textwidth}
        \centering
    \includegraphics[width=.9\linewidth]{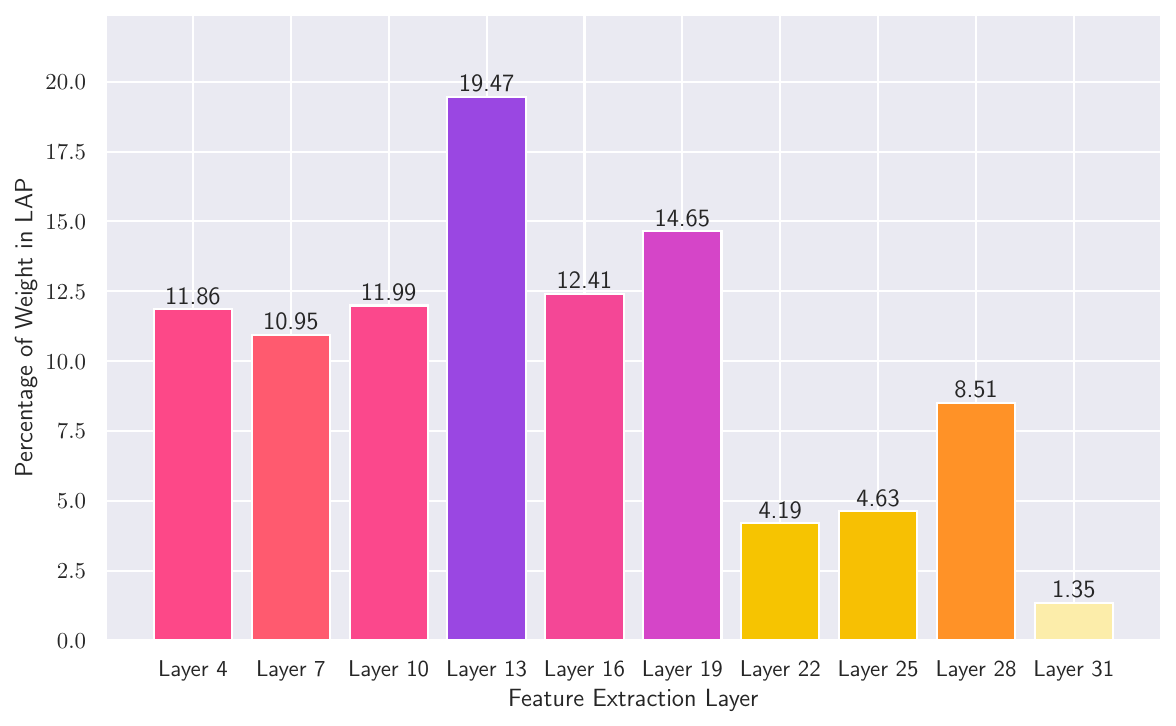}
    \caption{An LAP using only every 3rd layer of the encoder VLM learns more uniform layer weights. }
     \label{fig:layer_importance_revised}
    \end{minipage}
\end{figure}

We further discovered that LAP often pools the information from a contiguous set of layers to form the final representation for a single token.
When plotting the Query-Key norms for individual tokens (Fig.~\ref{fig:local_clusters}), we found highly clustered activation patterns. For example, the word "drinking" shows clusters for the 1st-4th or 21st-24th layer. Activations of adjacent layers in the transformer are highly similar, as shown in Fig.~\ref{fig:layer_similarity} and in other works \cite{krause2025tread, lawson2025residual, liu2023dejavu}.
We argue that while the image generator still benefits from extracted representations from all depths of a VLM encoder, considering every layer adds unnecessary redundancy and suboptimal parameter utilization.

\begin{figure}[t]
    \centering
    \includegraphics[width=0.95\linewidth]{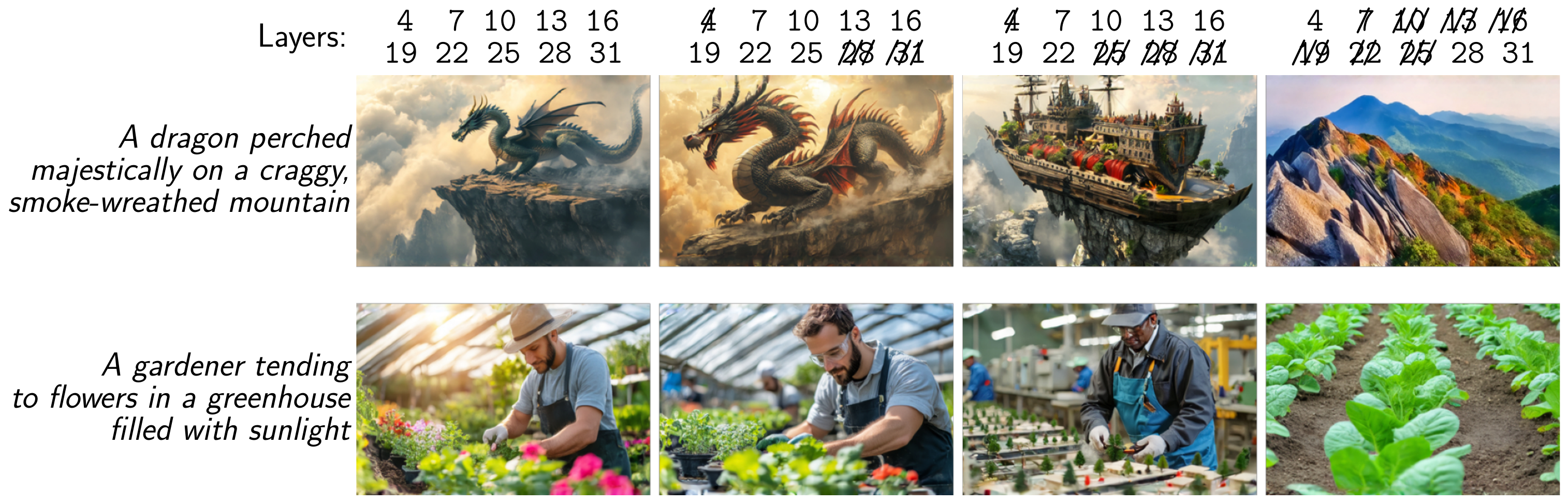}
    \caption{Qualitative analysis of different layer impact on final image. We drop crossed-out layers in LAP aggregation. Middle layers are crucial to capture the overall scene composition. In contrast first and last layers only capture rudimentary aspects of the scene. }
    \label{fig:layer_dropping}
\end{figure}

Based on these insights, our final LAP architecture takes in every third layer of the input encoder. 
This setting balances the capture of relevant information against computational overhead and eliminates local clusters. 
A model trained with this revised configuration now exhibits more uniform weight allocation as seen in Fig.~\ref{fig:layer_importance_revised}. 
Notably, the penultimate VLM layer contributes the least to the pooled representation, despite its prominent use in current methods. In the final \model~architecture, we perform a VLM layer dropout experiment as demonstrated in Fig.~\ref{fig:layer_dropping}. We observe that image generation does not strongly rely on the first and last layers. When zeroing out the respective weights during pooling, the overall image composition remains unchanged. In contrast, dropping information from the middle layers results in significant deviations in the output. 

\begin{tcolorbox}[leftrule=1.5mm,top=0.8mm,bottom=0.5mm,title=\model~LAP Layer Selection]
\begin{itemize}
    \item We extract features from every third layer across the depth of the VLM as the input for LAP
    \item This setup reduces overhead while maximizing information extraction from the VLM encoder 
\end{itemize}
\end{tcolorbox}

\subsection{Position Bias}\label{subsec:causual_attention}
In our initial analysis in Sec.~\ref{subsec:encoder_knowledge}, we identified cases where the model fails to accurately capture a key subject from the text prompt.
This issue can be attributed to bias introduced by \textit{causal attention} masking in the encoder transformer \cite{ma2024exploring}. 
Since a given token will only be attended to by the ones following it, information about a subject mentioned late in the prompt will be insufficiently represented.

We combat this bias by adding a simple refiner to the representation adapter. Similar to \citet{ma2024exploring}, this module consists of two standard transformers with full self-attention over the sequence length \texttt{sl}.
We found that this small bi-directional refiner significantly boosts performance over crude hidden-state extraction (See \citet{ma2024exploring} for detailed ablations). 
Consequently, our final \model~adapter combines Layerwise Attention Pooling with a two-block refiner, operating on the pooled representation. We find that both components are crucial in achieving optimal performance. 
We provide more details in App.~\ref{app:causual_attention}. 
These results further support our findings from Sec .~\ref{subsec:encoder_knowledge} that LAP remains the superior extraction approach.

\begin{tcolorbox}[leftrule=1.5mm,top=0.8mm,bottom=0.5mm,title=\model~Refiner]
\begin{itemize}
    \item The final \model~adaptor combines Layerwise Attention Pooling with a bi-directional refiner
    \item This refiner adds two transformer blocks of full self-attention on the aggregated sequence to mitigate position biases
\end{itemize}
\end{tcolorbox}

\subsection{VLM-Enabled Rewriting Injection with Flexible Inference (\rewrite)}

Next, we consider in more detail how to best leverage the VLM's inherent capabilities for conditioning the DiT. 
We propose VLM-Enabled Rewriting Injection with Flexible Inference (\rewrite), which folds the VLM's world knowledge and reasoning into our unified representation space.


%

Given a user input consisting of text and images \texttt{prompt, local image patches, image thumbnail}, we use a dedicated \texttt{system prompt} to instruct the VLM to generate a \texttt{target prompt} as a detailed description of the intended image. 
Prompt rewriting has become a common practice for providing dense, detail-rich instructions to generative image models \cite{openai2023dalle3, liu2024playgroundv3improvingtexttoimage, esser2024scalingrectifiedflowtransformers, brack2025traintexttoimagemodelevaluating}. However, rewriting prompts in our setting is fundamentally different from these approaches and offers some additional benefits. 

Firstly, \rewrite~does not require a standalone rewriter with subsequent input encoding of the adjusted prompt. Instead, we perform a single forward pass without re-encoding. Thus, the target tokens will still attend to the original prompt and retain that context in the final representations. For multi-modal inputs, the context provided through attention produces aligned features between modalities. 
Secondly, repetition of important information from the original prompt can further mitigate position attention biases (Sec.~\ref{subsec:causual_attention}). 
For text-to-image generation, \rewrite~ only uses rewritten tokens for DiT conditioning. In multimodal settings, we also inject all image tokens (patch and thumbnail) to ensure preservation of fine-grained details. 

\begin{figure}[t]
    \centering
    \begin{subfigure}[t]{0.6\textwidth}
        \centering
        \includegraphics[height=12.5em]{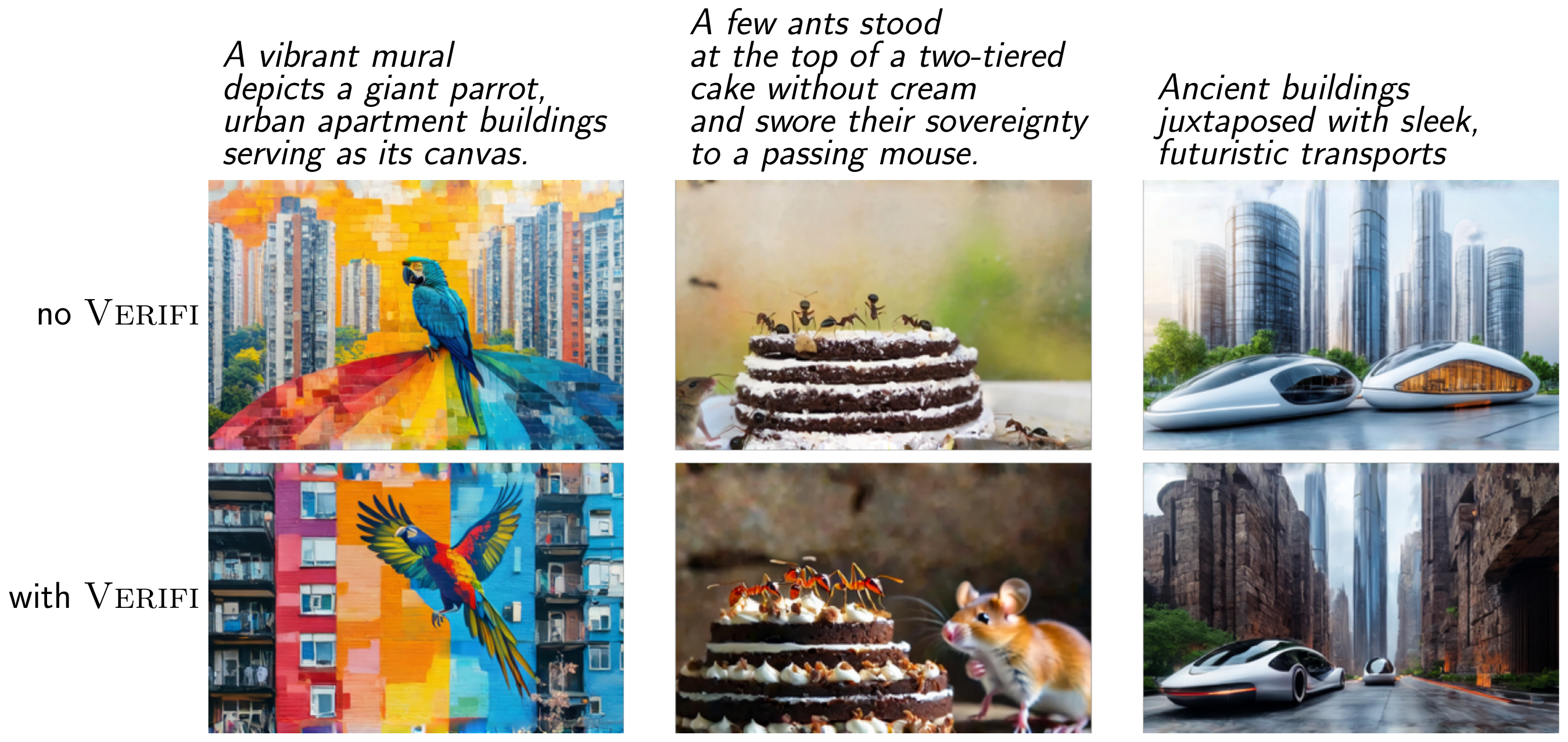}
        \caption{Qualitative Examples}
        \label{fig:self_rewrite_qual}
    \end{subfigure}%
    \hfill
    \begin{subfigure}[t]{0.38\textwidth}
        \centering
        \includegraphics[height=11.5em]{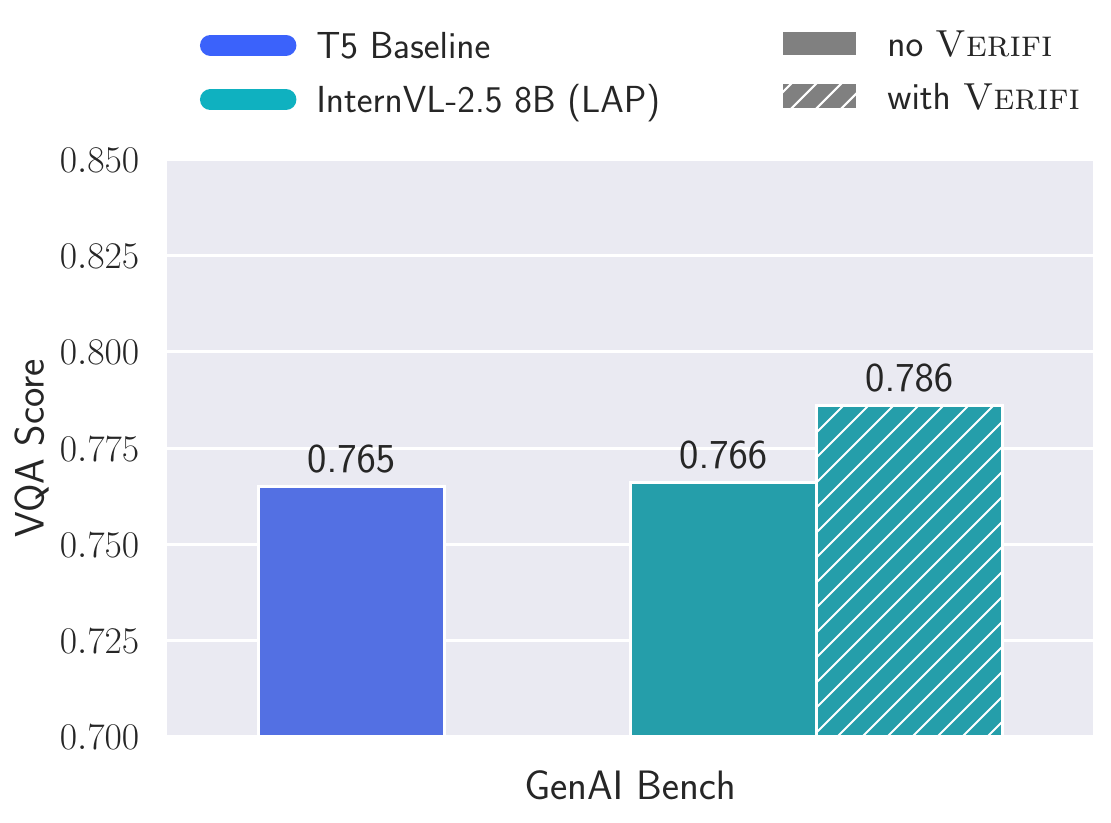}
        \caption{GenAI Bench Evaluation}
        \label{fig:self_rewrite_genai}
    \end{subfigure}
    \caption{\rewrite~improves prompt following. Comparison of InternVL-2.5-8B conditioned DiT with and without \rewrite. Especially, complex prompts involving multiple subjects are generated more accurately.}
    \label{fig:self_rewrite}
\end{figure}

We depict qualitative results and benchmark evaluations for self-rewriting in Fig.~\ref{fig:self_rewrite}. In general, we found that \rewrite~significantly improves prompt following performance. In some cases, the differences between images are small. But crucially, it reliably mitigates catastrophic failure cases that miss important aspects of the prompt entirely. In the examples in Fig.~\ref{fig:self_rewrite}, \rewrite~correctly places the parrot on the buildings, adds the mouse to the generated image, and depicts ancient buildings.
Further, we observed that \rewrite~also improves performance on already long, detailed prompts. Consequently, the benefits extend beyond embellishing details and also contribute to mitigating position bias. 

While we use the same system prompt during training, we can meaningfully influence the model's behavior by changing the system prompt at inference. Since we keep the VLM frozen and use its standard chat template during rewriting, all original capabilities of the model remain intact. We found that the usage of the chat template, an instruction-tuned model that was optimized for, is crucial in extracting meaningful features. Since \rewrite~imitates a turn in a regular chat interaction, we remain in-distribution of the VLM. 

\rewrite~also enables zero-shot reasoning over complex inputs, which we explore in more detail in Sec.~\ref{sec:emergent}.

\

\begin{tcolorbox}[leftrule=1.5mm,top=0.8mm,bottom=0.5mm,title=\model~Image Input Encoding]
\begin{itemize}
    \item The VLM in \model~uses \rewrite~to generate the final text prompt
    \item We use all image tokens for DiT conditioning, but only rewritten text
\end{itemize}
\end{tcolorbox}

\subsection{Finetuning vs. Training from Scratch}\label{subsec:finetune}
By now, we have established clear theoretical and practical benefits of a unified encoder. However, training such a model from scratch comes with significant computational requirements. Consequently, many prior works have sought out more efficient approaches to add new input modalities and capabilities to existing models \cite{bellagente2023multifusion, ye2025altdiffusion, qin2023gluegen}. 

We conduct a controlled experiment comparing training a model with a VLM encoder from scratch against adopting a pre-existing T5 checkpoint. To that end, we took a model trained on T5 for 100k steps and then switched to multimodal conditioning using InternVL-2.5-8B LAP.  
We observed that roughly 10k steps are sufficient for the new conditioning setup to generate coherent images.

When controlling for the total number of training samples, we find little difference in performance between models trained from scratch and switching from T5 halfway through. Both the benchmark performance and qualitative capabilities, including self-rewrite, multimodal, and zero-shot editing (see Sec.~\ref{sec:emergent}), are preserved in the continued model (details in App.~\ref{app:finetune}).
These results allow us to conclude that continual pre-training with unified encoders from a pre-existing model is as valid as training from scratch. Unless there are additional changes to the training setup, 
adopting an existing checkpoint will save compute resources with no obvious drawbacks. 
Conversely, there is also no benefit in using T5 for early training steps, as any model trained with a \model~approach will quickly converge to better text-to-image performance while enabling additional use cases. 

\begin{tcolorbox}[leftrule=1.5mm,top=0.8mm,bottom=0.5mm,title=\model~Training Regime]
\begin{itemize}
    \item For better compute utilization, we train the final \model~model by adapting an early T5-conditioned checkpoint. This yields no performance difference from training from scratch, further enabling \model~to be used on any pretrained T5-conditioned models.
\end{itemize}
\end{tcolorbox}

\begin{figure}[]
    \centering
    \includegraphics[width=\linewidth]{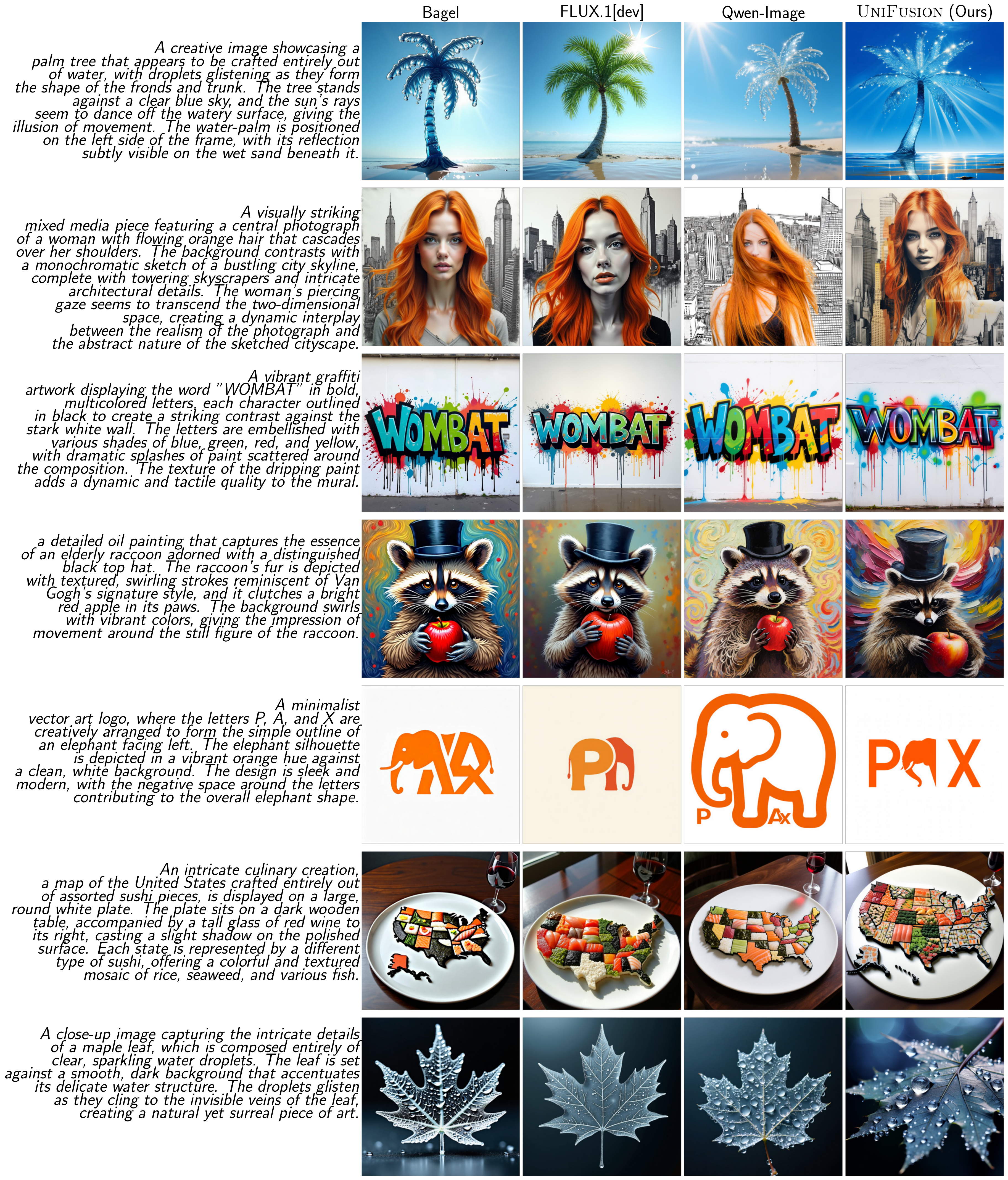}
    \caption{Qualitative comparison on long form text-to-image generation prompts comparing \model~to Bagel \cite{deng2025bagel}, Flux.1 [dev] \cite{flux2024} and Qwen-Image \cite{qwen-image}.}
    \label{fig:dpg_qualitative}
\end{figure}

\section{Final \model~Model}\label{sec:unifusion}
Finally, we integrate all the learnings from previous sections into a scaled-up model. We increase the DiT parameters to 8 billion and the total number of training samples to approximately 830 million. Instead of InternVL-2.5, we use InternVL3-8B \cite{zhu2025internvl3}. 

\subsection{\model~Design \& Training}
We design our final \model~model to extract features from every third layer of the VLM and aggregate them into a single representation via our LAP module. The LAP contains two transformer blocks aggregating the representation of any token across \textit{layers}. This sequence is then pooled into one dense representation with a simple fully connected layer.
The LAP is followed by a refiner of two bidirectional transformer blocks, mitigating position bias across the input \textit{sequence}. We inject the extracted representation only in the DiT's input sequence, which operates on a VAE latent space with a compression factor of $16$.

We only encode input images through the VLM and do not concatenate any additional VAE tokens to the DiT input. \model~leverages self-rewrite of user inputs, with only the image and rewritten tokens being used in DiT conditioning. For image inputs, we train the model on up to 10 tiles. 
Given our insights from Sec.~\ref{subsec:finetune}, we optimize compute usage by doing an early checkpoint handoff from a pre-existing T5 checkpoint. As described in Sec.~\ref{sec:architecture}, we train the base model on a mixture of text-to-image, image reconstruction, and joined text-and-image samples. Subsequently, we continue training with instruction data for image editing and reference workflows. We found roughly 10k steps of instruction training to be sufficient to support this task. 
For all stages of training, we use no web-scraped data and only rely on images with permissive licenses for generative image training. 
For this study, we do not perform any further post-training, which we leave for future work. 




\subsection{Evaluation}\label{subsec:unifusion_eval}

\paragraph{Qualitative Examples. }
We showcase text-to-image outputs of the \model~model in Figs.~\ref{fig:t2i_teaser} and~\ref{fig:teaser_t2ir}. Additionally, Figs.~\ref{fig:ie_teaser} and \ref{fig:zershot_multimage} depict image editing and reference examples. We use the same model for both image editing and text-to-image workflows.
Interestingly, we observed that continued training on image editing and image reference tasks also improved the model's text-to-image capabilities (see Sec.~\ref{sec:emergent}). 

\begin{figure}[t!]
    \centering
    \includegraphics[width=\linewidth]{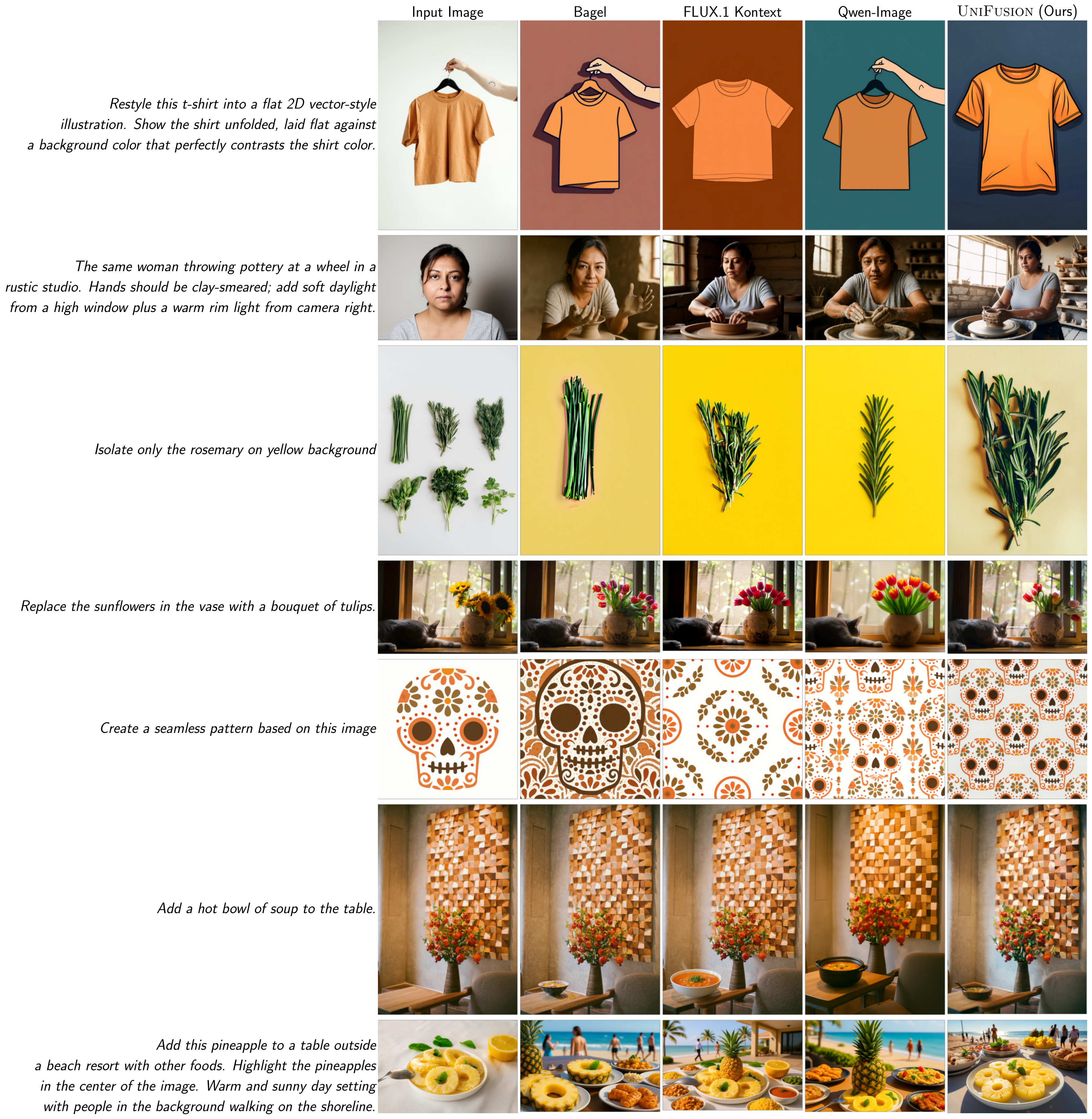}
    \caption{Qualitative comparison on image reference and editing tasks comparing \model~to Bagel \cite{deng2025bagel}, Flux.1 Kontext \cite{labs2025flux1kontext} and Qwen-Image \cite{qwen-image}. (Zoom in for details)}
    \label{fig:edit_comparison}
\end{figure}

In general, we find \model~to show strong performance for its size, efficient training regime, and without any supervised finetuning or reinforcement learning. \model~is capable of accurately generating images from long, complex prompts and excels in aesthetically pleasing, photorealistic generations, especially. 
%
In direct comparison with larger models, \model~remains highly competitive and especially benefits from improved visual understanding in image reference tasks. Importantly, for the direct text-to-image comparisons in Fig.~\ref{fig:dpg_qualitative} and image editing in Fig.~\ref{fig:edit_comparison}, Bagel and \model~are the only models using the same checkpoint for both tasks. In contrast, Flux and Qwen-Image rely on dedicated versions for each task. 

\paragraph{Quantitative Evaluation.}
Naturally, we ran several standardized benchmarks to judge the performance of the final model. 
Contrary, however, to under-trained ablations like those we conducted in Sec.~\ref{sec:ablations}, the usefulness of these benchmarks and metrics diminishes significantly when approaching their saturation. 

\begin{table}[t]
    \centering

\begin{tabular}{l rl rl rl rl rl}
\toprule
Category   & \multicolumn{2}{c}{Bagel \cite{deng2025bagel}} & \multicolumn{2}{c}{Flux.1 [dev] \cite{flux2024}} & \multicolumn{2}{c}{Qwen-Image \cite{qwen-image}} & \multicolumn{2}{c}{\model~(Ours)} \\
& Avg. & Top-4 & Avg. & Top-4 & Avg. & Top-4 & Avg. & Top-4 \\
\midrule
\textbf{Macro Avg} & {0.715}$\phantom{\bullet}$ & {0.901}$\phantom{\bullet}$ & {0.693}$\phantom{\bullet}$ & {0.899}$\phantom{\bullet}$ & \textbf{0.802}$\bullet$ & \textbf{0.943}$\bullet$ & \textbf{0.731}$\circ$ & \textbf{0.915}$\circ$\\
\textbf{Micro Avg} &  {0.786}$\phantom{\bullet}$ & {0.873}$\phantom{\bullet}$ & {0.753}$\phantom{\bullet}$ & {0.851}$\phantom{\bullet}$ & \textbf{0.841}$\bullet$ & \textbf{0.914}$\bullet$ & \textbf{0.787}$\circ$ & \textbf{0.880}$\circ$  \\
 \hline
entity - whole & \textbf{0.904}$\circ$ & {0.988}$\phantom{\bullet}$ & {0.880}$\phantom{\bullet}$ & {0.984}$\phantom{\bullet}$ & \textbf{0.942}$\bullet$ & \textbf{0.995}$\bullet$ & {0.894}$\phantom{\bullet}$ & \textbf{0.989}$\circ$ \\
entity - part & \textbf{0.814}$\circ$ & {0.929}$\phantom{\bullet}$ & {0.785}$\phantom{\bullet}$ & {0.924}$\phantom{\bullet}$ & \textbf{0.869}$\bullet$ & \textbf{0.969}$\bullet$ & {0.805}$\phantom{\bullet}$ & \textbf{0.938}$\circ$ \\
entity - state & {0.667}$\phantom{\bullet}$ & {0.925}$\phantom{\bullet}$ & {0.617}$\phantom{\bullet}$ & {0.890}$\phantom{\bullet}$ & \textbf{0.733}$\bullet$ & \textbf{0.940}$\circ$ & \textbf{0.673}$\circ$ & \textbf{0.943}$\bullet$ \\
attribute - color & \textbf{0.823}$\circ$ & {0.962}$\phantom{\bullet}$ & {0.779}$\phantom{\bullet}$ & {0.958}$\phantom{\bullet}$ & \textbf{0.866}$\bullet$ & \textbf{0.984}$\bullet$ & {0.803}$\phantom{\bullet}$ & \textbf{0.963}$\circ$ \\
attribute - size & {0.740}$\phantom{\bullet}$ & {0.904}$\phantom{\bullet}$ & {0.730}$\phantom{\bullet}$ & {0.882}$\phantom{\bullet}$ & \textbf{0.795}$\bullet$ & \textbf{0.914}$\circ$ & \textbf{0.771}$\circ$ & \textbf{0.930}$\bullet$ \\
attribute - shape & \textbf{0.702}$\circ$ & {0.873}$\phantom{\bullet}$ & {0.662}$\phantom{\bullet}$ & {0.873}$\phantom{\bullet}$ & \textbf{0.777}$\bullet$ & \textbf{0.919}$\bullet$ & {0.682}$\phantom{\bullet}$ & \textbf{0.879}$\circ$ \\
attribute - texture & {0.703}$\phantom{\bullet}$ & {0.922}$\phantom{\bullet}$ & {0.647}$\phantom{\bullet}$ & {0.900}$\phantom{\bullet}$ & \textbf{0.779}$\bullet$ & \textbf{0.945}$\bullet$ & \textbf{0.729}$\circ$ & \textbf{0.926}$\circ$ \\
attribute - other & {0.652}$\phantom{\bullet}$ & {0.891}$\phantom{\bullet}$ & {0.625}$\phantom{\bullet}$ & {0.894}$\phantom{\bullet}$ & \textbf{0.720}$\bullet$ & \textbf{0.928}$\phantom{\bullet}$ & \textbf{0.698}$\circ$ & \textbf{0.931}$\bullet$ \\
relation - spatial & {0.706}$\phantom{\bullet}$ & {0.942}$\phantom{\bullet}$ & {0.677}$\phantom{\bullet}$ & {0.946}$\phantom{\bullet}$ & \textbf{0.778}$\bullet$ & \textbf{0.967}$\bullet$ & \textbf{0.712}$\circ$ & \textbf{0.947}$\circ$ \\
relation - non-spatial & {0.579}$\phantom{\bullet}$ & {0.807}$\phantom{\bullet}$ & {0.549}$\phantom{\bullet}$ & {0.761}$\phantom{\bullet}$ & \textbf{0.701}$\bullet$ & \textbf{0.890}$\bullet$ & \textbf{0.643}$\circ$ & \textbf{0.826}$\circ$ \\
global - & {0.639}$\phantom{\bullet}$ & {0.864}$\phantom{\bullet}$ & {0.641}$\phantom{\bullet}$ & \textbf{0.897}$\circ$ & \textbf{0.714}$\bullet$ & {0.888}$\phantom{\bullet}$ & \textbf{0.688}$\circ$ & \textbf{0.898}$\bullet$ \\
other - count & {0.769}$\phantom{\bullet}$ & {0.933}$\phantom{\bullet}$ & {0.765}$\phantom{\bullet}$ & {0.944}$\phantom{\bullet}$ & \textbf{0.850}$\bullet$ & \textbf{0.961}$\bullet$ & \textbf{0.791}$\circ$ & \textbf{0.955}$\circ$ \\
other - text & {0.600}$\phantom{\bullet}$ & {0.771}$\phantom{\bullet}$ & \textbf{0.655}$\circ$ & \textbf{0.833}$\circ$ & \textbf{0.900}$\bullet$ & \textbf{0.958}$\bullet$ & {0.615}$\phantom{\bullet}$ & {0.771}$\phantom{\bullet}$ \\
\hline
Model Size &  \multicolumn{2}{c}{14B MoT} & \multicolumn{2}{c}{12B} & \multicolumn{2}{c}{20B} & \multicolumn{2}{c}{8B} \phantom{djadf} \\
\bottomrule
\end{tabular}
\caption{\model~achieves competitive performance against much larger models trained on more data. Scores on modified DPG-Bench. We report average and best generation across four seeds at 1024px resolution. Macro Average is taken as the mean over scores per category, whereas Micro averages scores across all prompts. Results are scored by Gemma-3-27B with extensive CoT to reduce hallucinations in scoring. $\bullet$ and $\circ$ denote best and second-best score, respectively.}
\label{tab:dpg_metric}
\end{table}

For example, consider GenEval \cite{gosh2023geneval} and DPG-Bench \cite{hu2024ella}, two popular benchmarks for evaluating prompt-following capabilities. 
For both, we found the originally proposed evaluation settings to have immensely high error rates in accurately judging generated images. The noise of the benchmark itself far exceeded performance differences between models of a few percentage points. We also observed other crucial issues, such as unintentional penalization of photorealistic outputs, questions that cannot be answered objectively, or inaccurate score aggregation in DPG-Bench\footnote{Surprisingly, this implementation error appears to have gone largely unnoticed, despite being documented as an issue in the official GitHub repository. This bug has led to multiple papers reporting mathematically impossible results. For example, in the Qwen-Image paper, all category scores are reported to be higher than the overall average (Tab.~3, Page 21 \cite{qwen-image}).}. We provide more details in App.~\ref{app:benchmark_reliability}.

In Tab.~\ref{tab:dpg_metric}, we report \model's performance in comparison to other models on a revised version of DPG-Bench (See App.~\ref{app:dpg_refined}). Since most generative image applications provide users with up to four outputs, we report both the best-out-of-4 score and the average. 
Despite no post-training and a limited training sample, \model~remains competitive with significantly larger, heavily post-trained models. The qualitative comparison in Fig.~\ref{fig:dpg_qualitative} also highlights the competitive prompt-following capabilities of \model. Further we found, \model~less likely to generate characteristic AI artifacts like over-saturated colors and smoothed textures. 

When directly comparing the scores of Qwen-Image and \model, we see a larger gap between the Avg. and best-out-4 scores for \model. The difference in Macro Avg. scores is $0.142$ and $0.184$ for Qwen-Image and \model, respectively. We believe this to be a direct result of the post-training for Qwen-Image.

\begin{figure}[t]
    \centering
    \includegraphics[width=.95\linewidth]{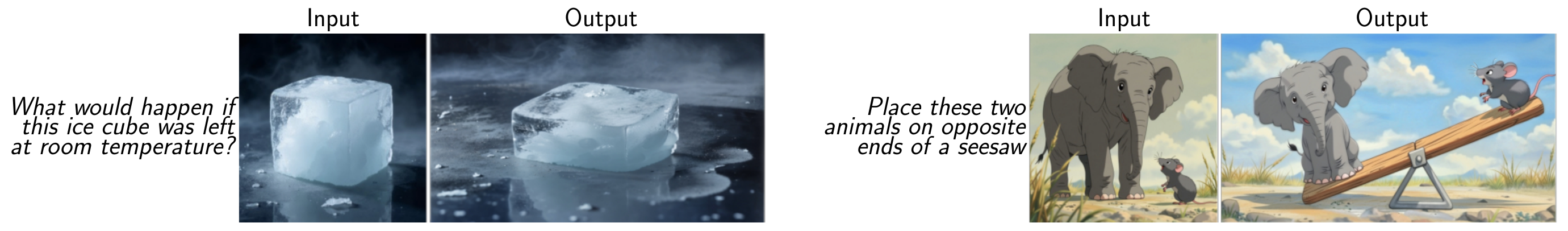}
    \caption{The unified VLM encoder enables advanced visual reasoning for textual image editing. (Examples on early checkpoint and not indicative of final model quality)  
    }
    \label{fig:visual_reasoning}
\end{figure}

\begin{figure}[b!]
    \centering
    \begin{subfigure}[t]{\textwidth}
        \centering
        \includegraphics[width=.9\linewidth]{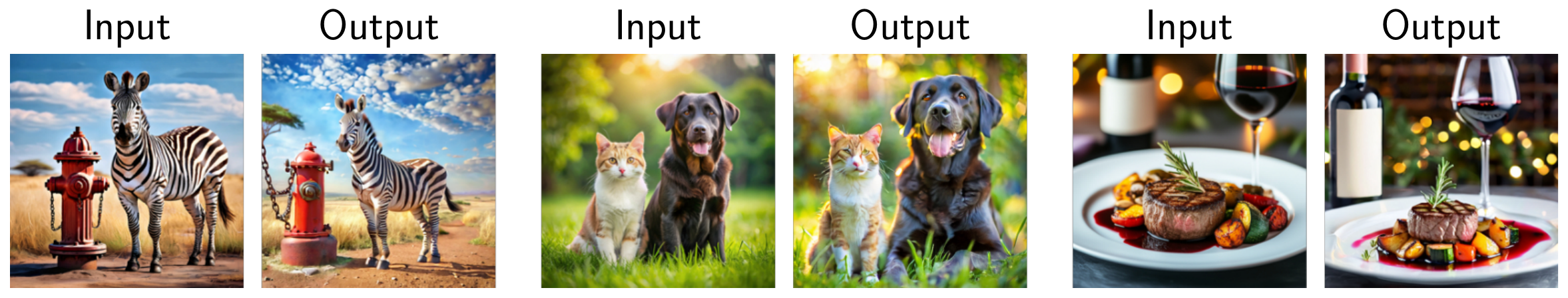}
        \caption{Zero-shot image-to-image generation. Examples generated by a model only trained on text-to-image generation. When presented with image features, the model captures overall scene composition and a high level of detail.  }
        \label{fig:zeroshot_i2i}
    \end{subfigure}%
    \hfill
    \begin{subfigure}[t]{\textwidth}
        \centering
        \includegraphics[width=.9\linewidth]{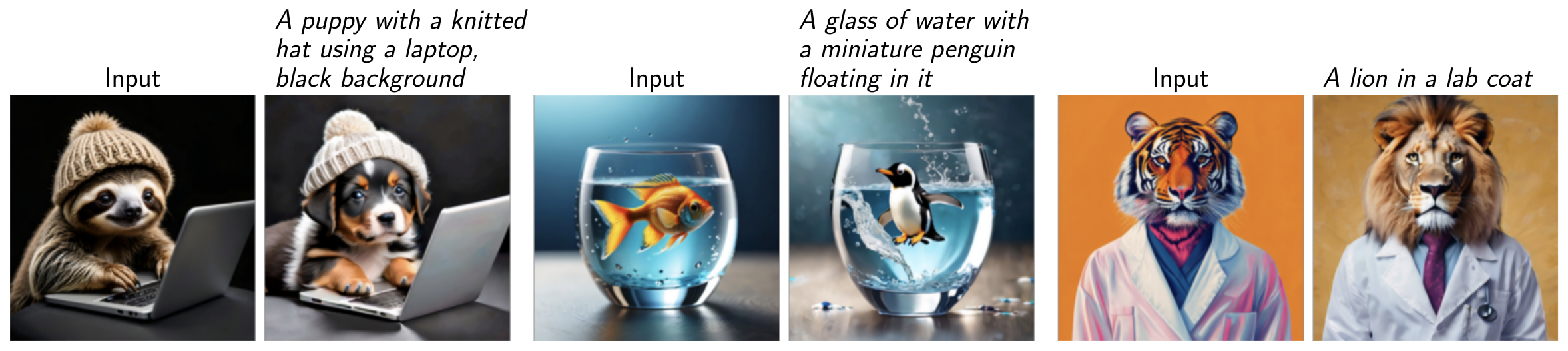}
        \caption{Zero-shot image editing. Examples generated by a model never trained on image editing.  }
        \label{fig:zeroshot_editing}
    \end{subfigure}
    \hfill
    \caption{Conditioning image generation on LAP extracted VLM features enables zero-shot generalisation to unseen tasks and modalities. }
    \label{fig:layer_selection}
\end{figure}
\section{Emergent Abilities}\label{sec:emergent}

During our experiments, we observe \model~to exhibit many valuable zero-shot capabilities without explicitly being trained for them. 
This behavior is a direct benefit of a unified VLM encoder architecture.
Any of the capabilities learned from the VLM's extensive training regime are retained and transferred to image generation tasks. Additionally, the unified space of contextualized text and image eliminates large distribution shifts between tasks. 

\subsection{Reasoning \& Complex Prompts}
\rewrite~allows the models to explicitly leverage the world knowledge and reasoning capabilities of the encoder VLM. In Fig.~\ref{fig:teaser_t2ir}, we show text-to-image examples using highly abstract text inputs. The model is capable of decomposing these instructions without any external sources. 


For example, given the prompt:
"The animal that represents the
zodiac sign between Aries and Gemini", InternVL rewrites "A majestic bull stands in a field of golden wheat, its horns curved in a fierce display of strength and virility". 
The VLM correctly references the zodiac sign Taurus, which is represented by a bull, and decodes the user input into a new prompt, allowing the DiT to successfully generate the animal. 

We observed similar capabilities for visual reasoning examples. For example, we can decompose hypothetical scenarios to perform image editing requiring multi-hop reasoning and world knowledge. We showcase some examples in Fig.~\ref{fig:visual_reasoning}. 
The VLM correctly reasons about hypothetical effects of the mass of different animals or impacts of temperature changes over time and decomposes those into an edit instruction. The DiT is then able to perform an edit satisfying the original user intent.

\begin{figure}[t]
    \centering
    \includegraphics[width=.99\linewidth]{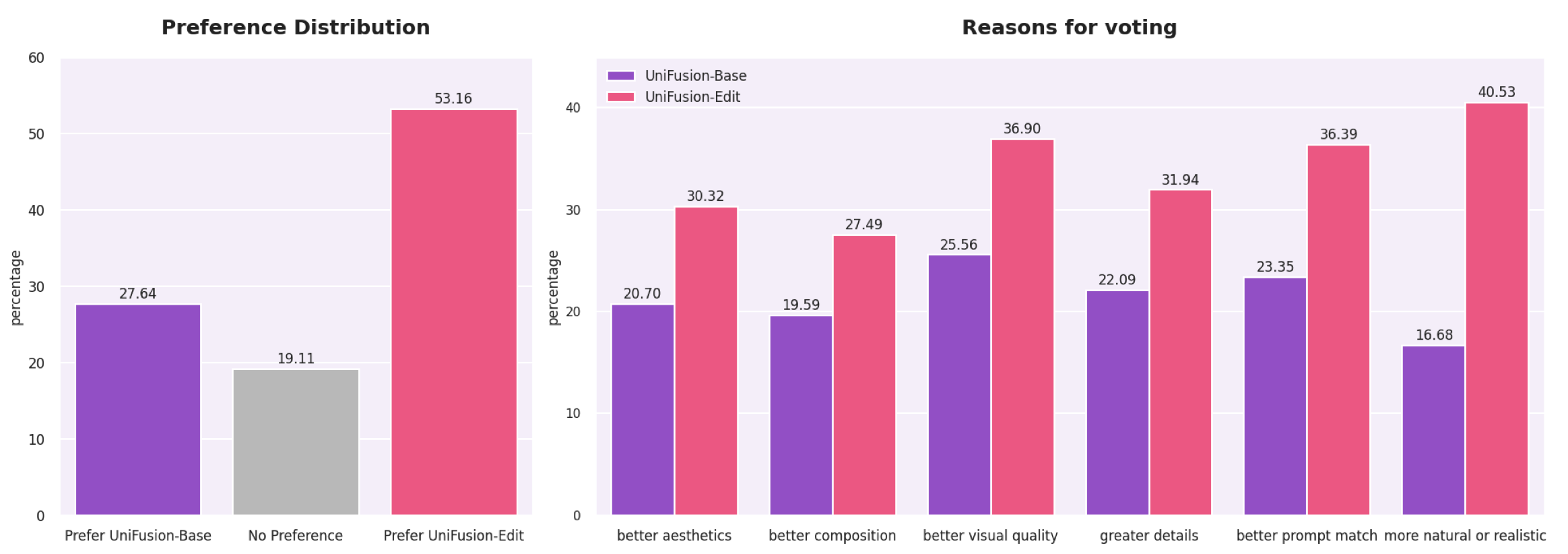}
    \caption{UniFusion-Edit leads UniFusion-Base by a significant margin in text-to-image A/B test with 180 annotators, 616 prompts across diverse concepts, with 2 seeds each (3 votes per image pair).  
    }
    \label{fig:capability_transfer_abtest}
\end{figure}

\subsection{Generalization to unseen modalities}
Throughout all experiments, we observed models to generalize well to inputs that were never observed during DiT training. Instead, these capabilities are a key benefit of a unified input space. 

For example, a model solely trained on text-to-image generation can still capture the semantics of image inputs as seen in Fig.~\ref{fig:zeroshot_i2i}. While the reconstruction is not pixel-perfect, the generated image still accurately captures the important aspects of the input image. This behavior can be attributed to the fact that the VLM yields decently aligned representation spaces for text and image, enabling zero-shot transfer to new modalities. 

Similarly, we find that models only trained on text \textit{or}
Image sequences, but not multimodal ones, can still be used for image editing. We show examples in Fig~\ref{fig:zeroshot_editing} where we successfully manipulate the content of an image by changing the respective textual scene description. 
Importantly, if a model is trained equally on text and image sequences, image tokens always take precedence at inference. This observation aligns with the findings of \citet{bellagente2023multifusion}. While they manipulated attention values directly to counteract this imbalance, we found that adjusting the training data composition works equally well. 
Specifically, when we condition the DiT with image tokens for the first 10-20\% of steps, and text tokens for the remaining 80-90\%, the output image preserves most of the input image's content and semantics, even though the model was never trained with any textual image editing data.

\begin{table}[t]
    \centering

\begin{tabular}{l rl rl rl rl rl}
\toprule
Category    & \multicolumn{2}{c}{\model-Base} & \multicolumn{2}{c}{\model-Edit}\\
& Avg. & Top-4 & Avg. & Top-4  \\
\midrule 
Macro Avg  &  \underline{0.699} & \underline{0.906} & \textbf{0.731} & \textbf{0.915} \\
Micro Avg & \underline{0.760} & \underline{0.863} & \textbf{0.787} & \textbf{0.880} \\ \hline
entity - whole & \underline{0.876} & \textbf{0.993} & \textbf{0.894} & \underline{0.989} \\
entity - part & \underline{0.772} & \textbf{0.938} & \textbf{0.805} & \textbf{0.938} \\
entity - state & \underline{0.627} & \underline{0.919} & \textbf{0.673} & \textbf{0.943} \\
attribute - other & \underline{0.661} & \underline{0.912} & \textbf{0.698} & \textbf{0.931} \\
attribute - color & \underline{0.785} & \textbf{0.970} & \textbf{0.803} & \underline{0.963} \\
attribute - size & \underline{0.744} & \underline{0.925} & \textbf{0.771} & \textbf{0.930} \\
attribute - shape & \underline{0.660} & \textbf{0.890} & \textbf{0.682} & \underline{0.879} \\
attribute - texture & \underline{0.704} & \underline{0.920} & \textbf{0.729} & \textbf{0.926} \\
relation - spatial & \underline{0.681} & \underline{0.943} & \textbf{0.712} & \textbf{0.947} \\
relation - non-spatial & \underline{0.591} & \textbf{0.826} & \textbf{0.643} & \textbf{0.826} \\
global - & \underline{0.670} & \underline{0.893} & \textbf{0.688} & \textbf{0.898} \\
other - count & \underline{0.752} & \textbf{0.961} & \textbf{0.791} & \underline{0.955} \\
other - text & \underline{0.560} & \underline{0.688} & \textbf{0.615} & \textbf{0.771} \\

\end{tabular}
\caption{Image Editing and Image Reference Training significantly improves \model~capabilities in text-to-image generation. Scores on modified DPG-Bench. We report average and best generation across four seeds at 1024px resolution. Macro Average is taken as the mean over scores per category, whereas Micro averages scores across all prompts. Results are scored by Gemma-3-27B with extensive CoT to reduce hallucinations in scoring. }
\label{tab:dpg_edit}
\end{table}

\subsection{Cross-Task Improvements}
In Section~\ref{sec:unifusion}, we observed that continued training on image editing and image reference tasks also improved the model's text-to-image quality.
In Tab.~\ref{tab:dpg_edit}, we see a significant improvement on DPG-Bench of over 2 percentage points in Micro Avg. We further conducted a human user study comparing checkpoints before and after training on editing data, as shown in Fig. \ref{fig:capability_transfer_abtest}.
Annotators strongly prefer the images generated by the \model-Edit checkpoint across all aspects of text-to-image generation. 
We hypothesize that this behavior is a direct benefit of a unified encoder architecture. Since the representation space always supported multimodal inputs, the transition from text-to-image towards editing is not a significant shift. Instead, this stage increases concept coverage and refines the model's representations. Since \model~eliminates the need to introduce VAE-encoded image reference inputs, the DiT does not need to adjust its embedding behavior. Consequently, we are now able to reap the benefits of further task coverage without the adverse effect of a new input structure.

\subsection{Zero-shot multi-reference capabilities}
Lastly, we also observed strong zero-shot abilities for image reference tasks. 
The editing data in Sec.~\ref{sec:unifusion}, contained only examples with a single reference image. Additionally, all training samples fix the input and output images to the same aspect ratio. 

Nonetheless, the examples in Fig.~\ref{fig:zershot_multimage} demonstrate that \model~is capable of accurately composing scenes from multiple reference images.
In these use cases \model~also seamlessly handles input and output images of different aspect ratios and resolutions and applies unprompted shifts in perspective when needed. 
For example, the scene reference on top of the pyramids is given in a different aspect ratio than the output image. \model~expands the scene and slightly shifts the perspective to account for that change while preserving fine-grained image details. 

\section{Conclusion}

\paragraph{Limitations \& Discussion.} 
While our \model~approach provides significant benefits over other conditioning methods, there are some limitations worth discussing. 

Naturally, auto-regressive self-rewriting of all input prompts with an 8B transformer comes with an increase in compute and runtime during encoding. However, given the prominence of prompt rewriting in general and other approaches using VLM conditioning, this limitation is not unique to \model. 

Furthermore, we identified some issues related to rendering text in scenes, which also impact the respective scores in Tab.~\ref{tab:dpg_metric}. In general, the model is capable of generating and editing typography, as shown in Figs.~\ref{fig:t2i_teaser} and \ref{fig:ie_teaser}. However, we found InternVL to be particularly bad at spelling. Consequently, the model often misspells text in the rewritten prompt, leading to the generation of incorrect or illegible text. 
We can further pinpoint this issue to InternVL specifically by having an external model perform the rewriting. In this scenario, even when re-encoding the text through InternVl, \model~reliably generates text in images. 

So far, our experiments have focused on InternVL as a candidate encoder. To ensure that \model~generalises beyond one VLM family, we trained an additional model based on Gemma. Overall, we found Gemma-based models to work similarly well and conclude that \model~is not limited to any specific VLM. We share more details on the Gemma experiments in App.~\ref{app:gemma}.

In conclusion, this work introduced \model, a framework that uses a single Vision-Language Model (VLM) as a unified encoder for generative image models. 

We proposed a novel Layerwise Attention Pooling (LAP) module, which aggregates features from multiple layers of a frozen VLM. Through structured experiments, we demonstrated that LAP outperforms other architectures, such as last-layer encoding and key-value fusion, in both prompt adherence and the preservation of fine-grained image details. Additionally, we provide strong evidence for critical design choices in best leveraging VLMs for generative image tasks. We derive practical suggestions on layer selection, bi-directional refiners, and the benefits of \rewrite. The \model~approach successfully eliminates the need for multiple image encoders.

By leveraging the powerful reasoning and world knowledge of the VLM, \model~gains significant zero-shot capabilities and generalises well to unseen use cases. The model can interpret complex, abstract prompts and perform visual reasoning and image reference tasks without explicit training. Furthermore, this framework allows for efficient adaptation of existing models, making it a computationally viable approach.
Overall, our findings establish that the \model~approach is a robust and flexible strategy to use VLMs as unified encoders. This research paves the way for developing more capable and intuitive image generation systems. 

\section*{Acknowledgments}
We would like to thank Alexandru Costin and Jingwan Lu for their support. 
We thank Sai Bi for continued discussions, especially on the Llama series of ablations. Thank you to Mingze Xu, Felix Friedrich, Melissa Hall, and Rena Ju for their feedback on the initial draft of the paper. 
{
    \small
    \bibliographystyle{acl_natbib}
    \bibliography{main}
}

\clearpage

\maketitlesupplementary
\setcounter{section}{0}
\renewcommand{\thesection}{\Alph{section}}
\begin{figure}[t!]
    \centering
    \includegraphics[width=0.9\linewidth]{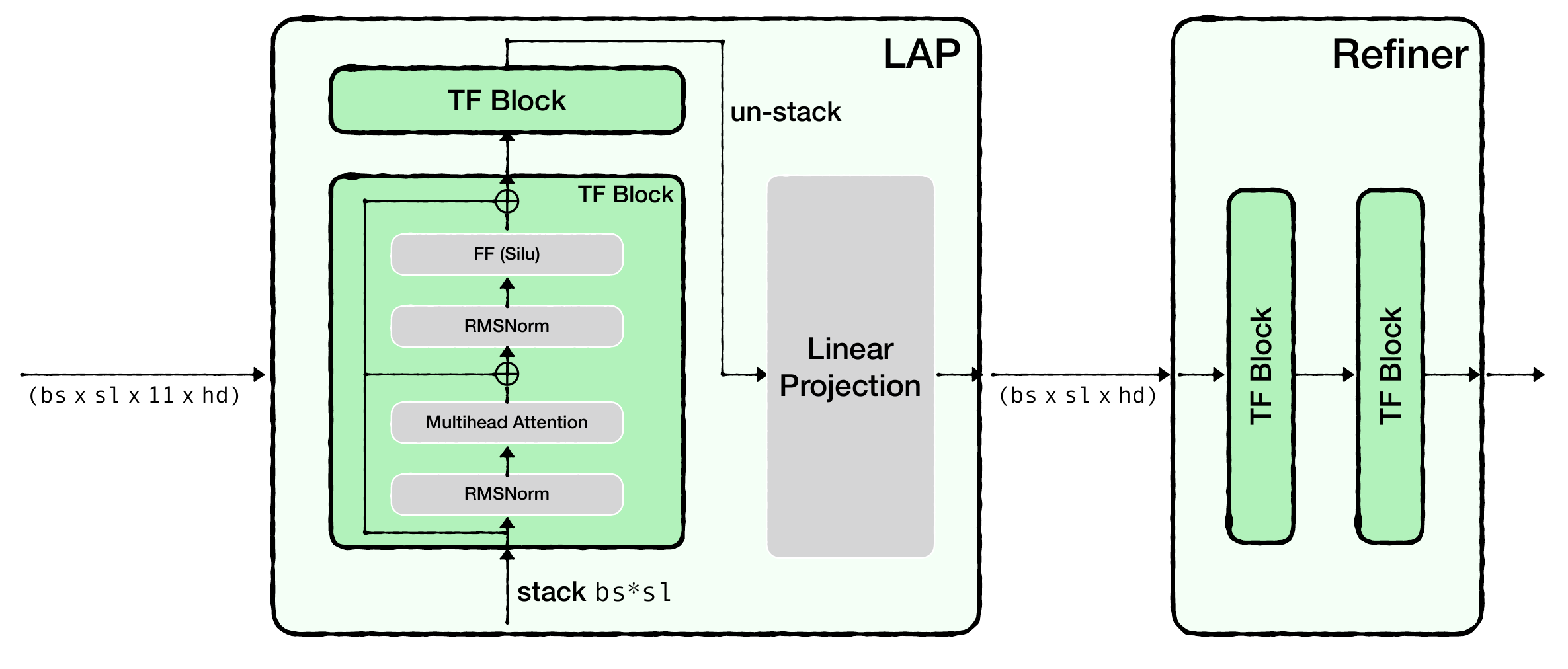}
    \caption{\model~adapter for layerwise representation aggregation. Representations from multiple VLM layers are aggregated using a Layerwise Attention Pooling (LAP). The aggregated representations are subsequently passed through a Refiner to mitigate position bias.}
    \label{fig:unifusion_architecture}
\end{figure}

\section{Additional Results \& Experimental Details}
In this section, we provide additional experimental details and results. 
\subsection{\model~layerwise representation aggregation}
We provide a visual aid for \model's representation aggregation adapter in Fig.~\ref{fig:unifusion_architecture}.
As mentioned in Sec.~\ref{sec:unifusion}, \model~extracts features from every third layer of the VLM and aggregates them into a single representation via our LAP module. The LAP contains two transformer blocks aggregating the representation of any token across \textit{layers}. This sequence is then pooled into one dense representation with a simple fully connected layer.
The pooled LAP representation is followed by a refiner of two bidirectional transformer blocks, mitigating position bias across the input \textit{sequence}.

In this context, our transformer blocks use multi-head attention with 32 attention heads. We apply RMS normalization before and after self-attention. These operations are followed by a feed-forward block using Silu as the activation function. We expand and contract the hidden dimension by a factor of $1.3$ for the non-linear activation.

\begin{figure}[t]
    \centering
    \begin{subfigure}[t]{0.6\textwidth}
        \centering
        \includegraphics[height=12.5em]{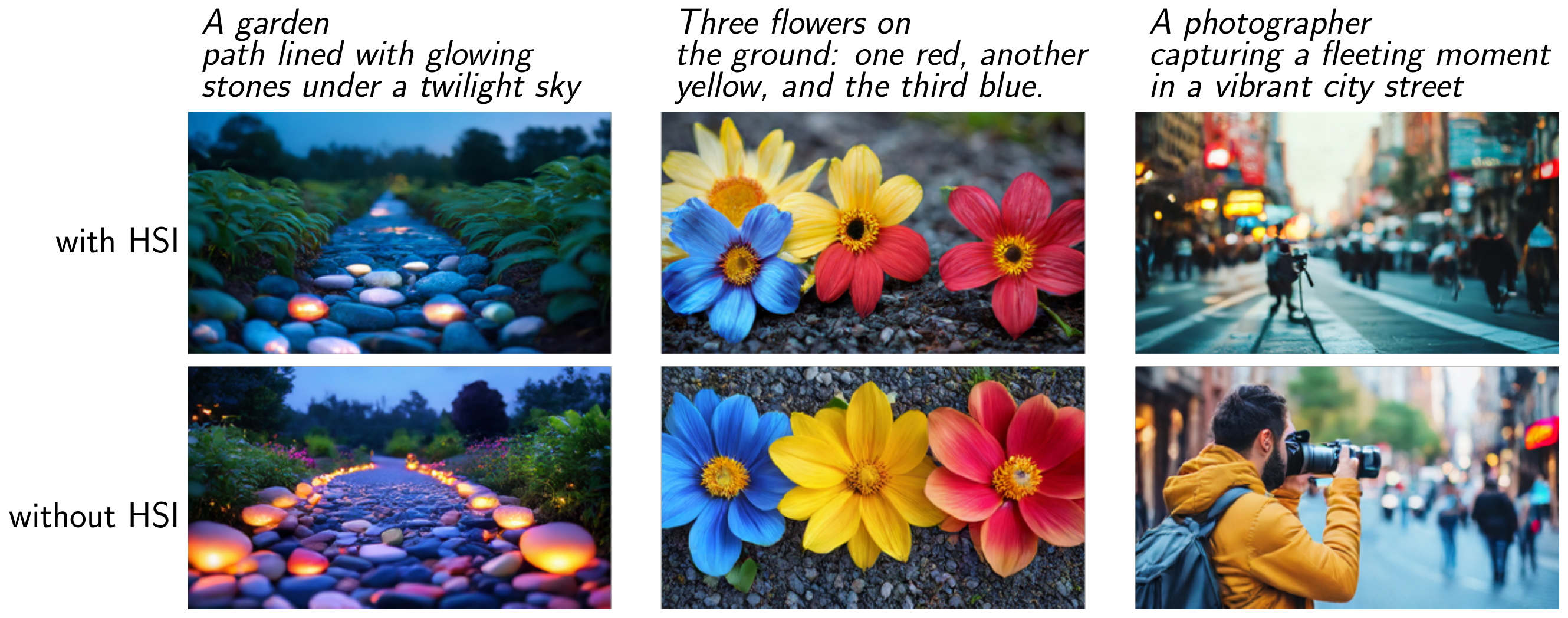}
        \caption{Qualitative Examples}
        \label{fig:injection_qual}
    \end{subfigure}%
    \hfill
    \begin{subfigure}[t]{0.38\textwidth}
        \centering
        \includegraphics[height=12.5em]{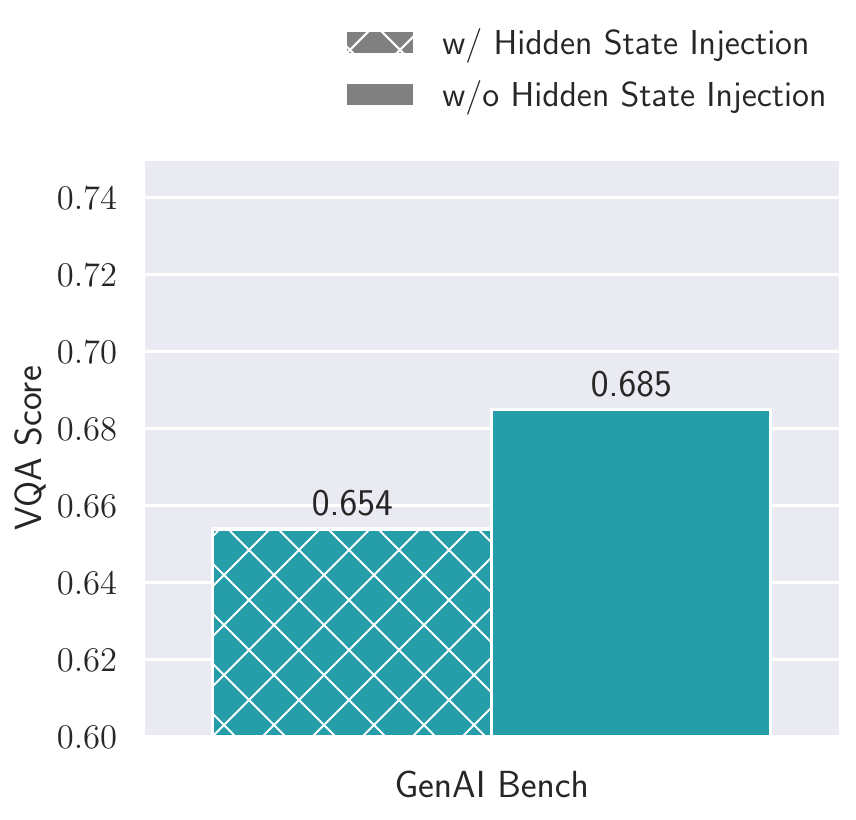}
        \caption{GenAI Bench Evaluation}
        \label{fig:injection_genai}
    \end{subfigure}
    \caption{Injecting aggregated representations at different DiT depth does not improve performance. Comparison of InternVL-2.5-8B with LAP feature extraction. First version injects dedicated representations per DiT layer (with HSI), second version pools one representation for DiT conditioning (without HSI). Comparison at 200k training steps}
    \label{fig:injection}
\end{figure}

\begin{figure}[b]
    \centering
    \begin{subfigure}[t]{0.6\textwidth}
        \centering
        \includegraphics[height=13em]{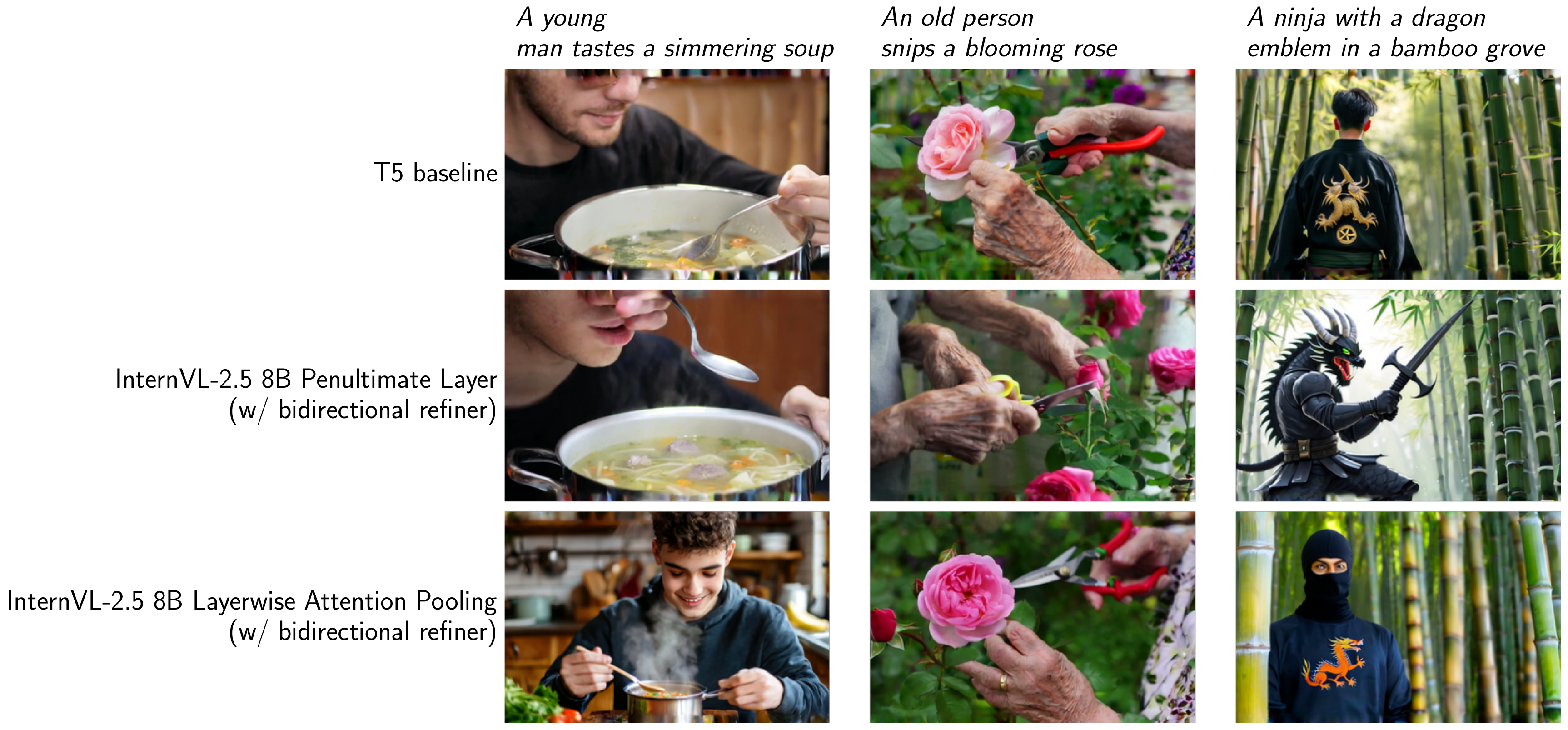}
        \caption{Qualitative Examples}
        \label{fig:refiner_qual}
    \end{subfigure}%
    \hfill
    \begin{subfigure}[t]{0.35\textwidth}
        \centering
        \includegraphics[height=12.5em]{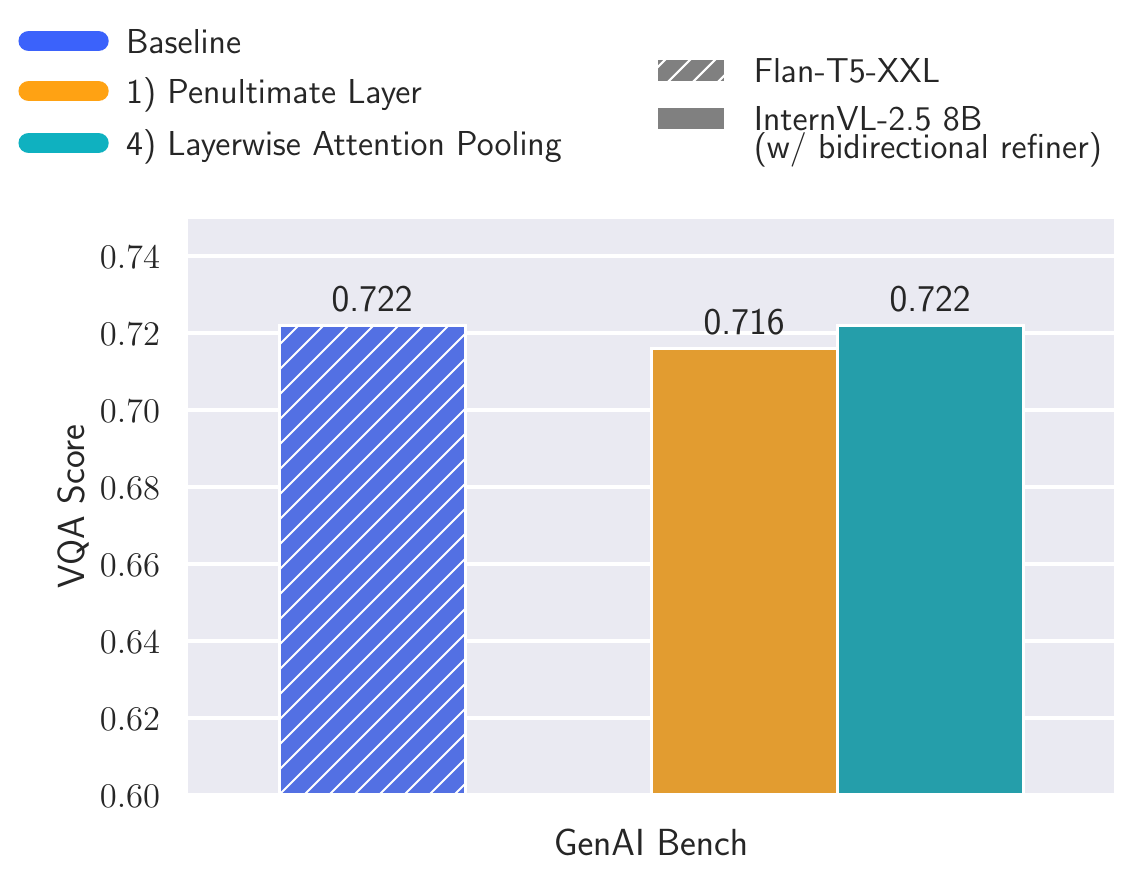}
        \caption{GenAI Bench Evaluation}
        \label{fig:refiner_genai}
    \end{subfigure}
    \caption{Evaluation of bi-directional refiner impact. InternVL2.5-8B model with refiner closes the performance gap to the T5 baseline (cf.~Fig.~\ref{fig:text_to_image_ablations}. Nonetheless, layer-wise attention pooling still outperforms representation extraction from the last layers. Comparison at 250k training steps.}
    \label{fig:refiner}
\end{figure}

\begin{figure}[t]
    \centering
    \begin{subfigure}[t]{0.6\textwidth}
        \centering
        \includegraphics[height=13em]{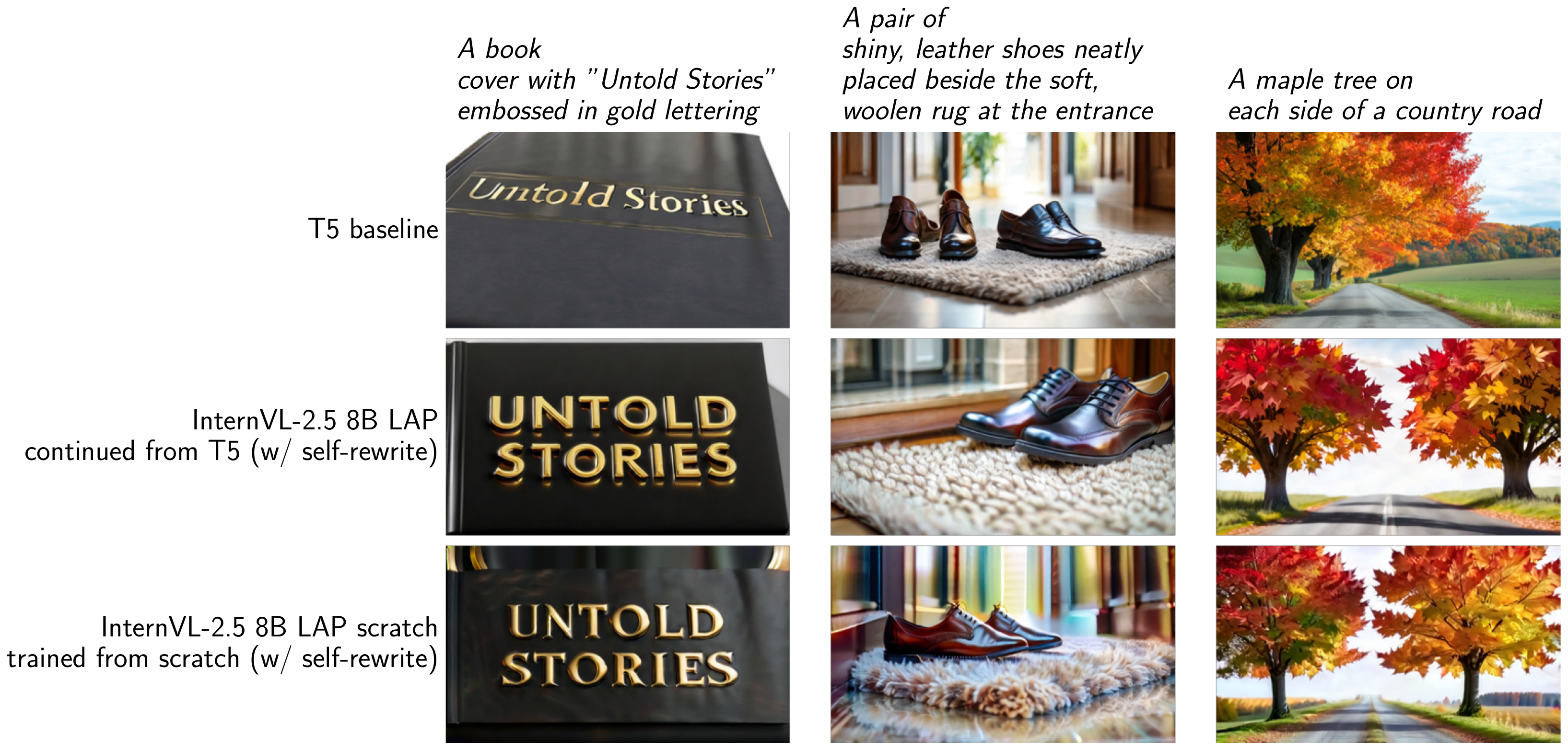}
        \caption{Qualitative Examples}
        \label{fig:rcontinue_vs_scratch_qual}
    \end{subfigure}%
    \hfill
    \begin{subfigure}[t]{0.35\textwidth}
        \centering
        \includegraphics[height=12.5em]{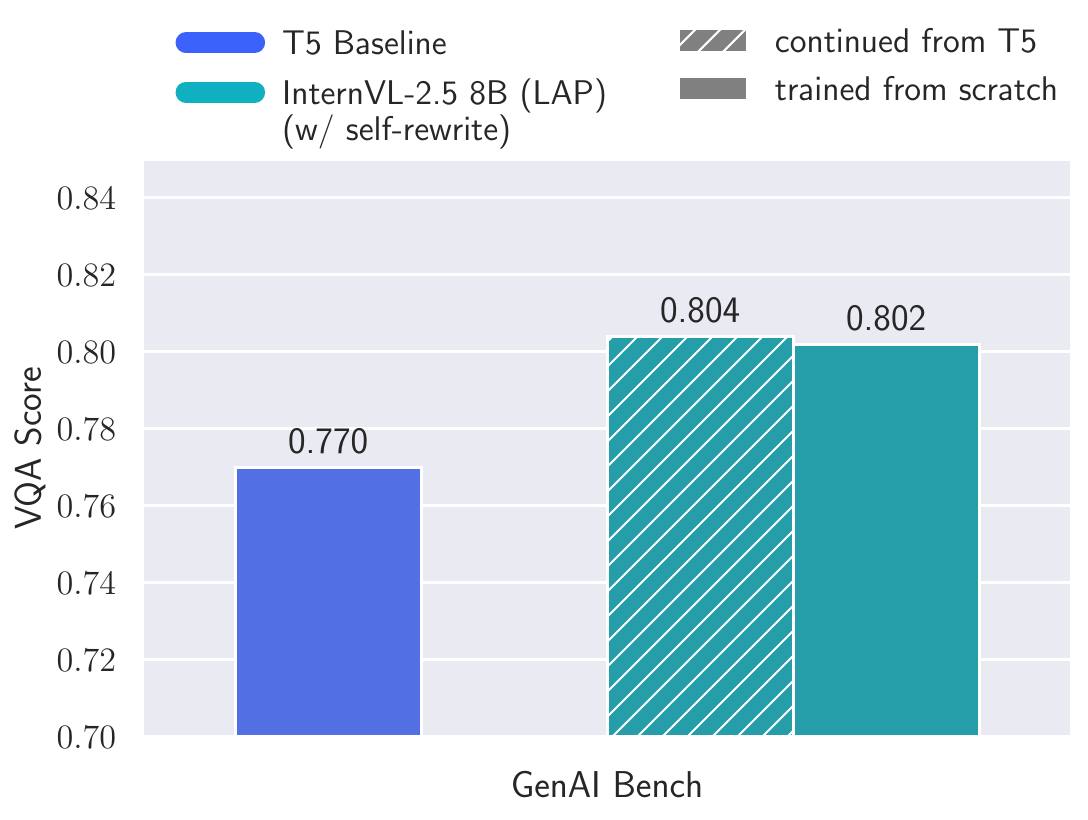}
        \caption{GenAI Bench Evaluation}
        \label{fig:rcontinue_vs_scratch_genai}
    \end{subfigure}
    \caption{Evaluation of training \model~conditioning from scratch vs. continuing from T5. Both approaches produce models with the same capabilities. Comparison at 250k training steps. Continued checkpoint switches from T5 to InternVL2.5-8B at 100k steps.}
    \label{fig:continue_vs_scratch}
\end{figure}

\begin{figure}[b!]
    \centering
    \begin{subfigure}[t]{\textwidth}
            \centering
            \includegraphics[width=.9\linewidth]{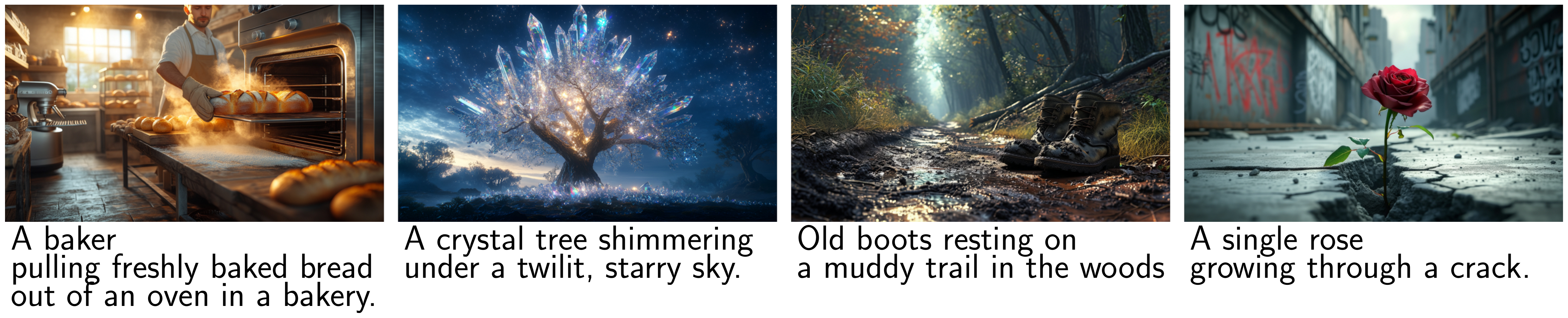}
            \caption{Text-to-image examples generated with \model-Gemma using self-rewrite. }
            \label{fig:gemma_t2i}
        \end{subfigure}%
        \vspace{15px}
        \begin{subfigure}[t]{\textwidth}
        \centering
        \includegraphics[width=.9\linewidth]{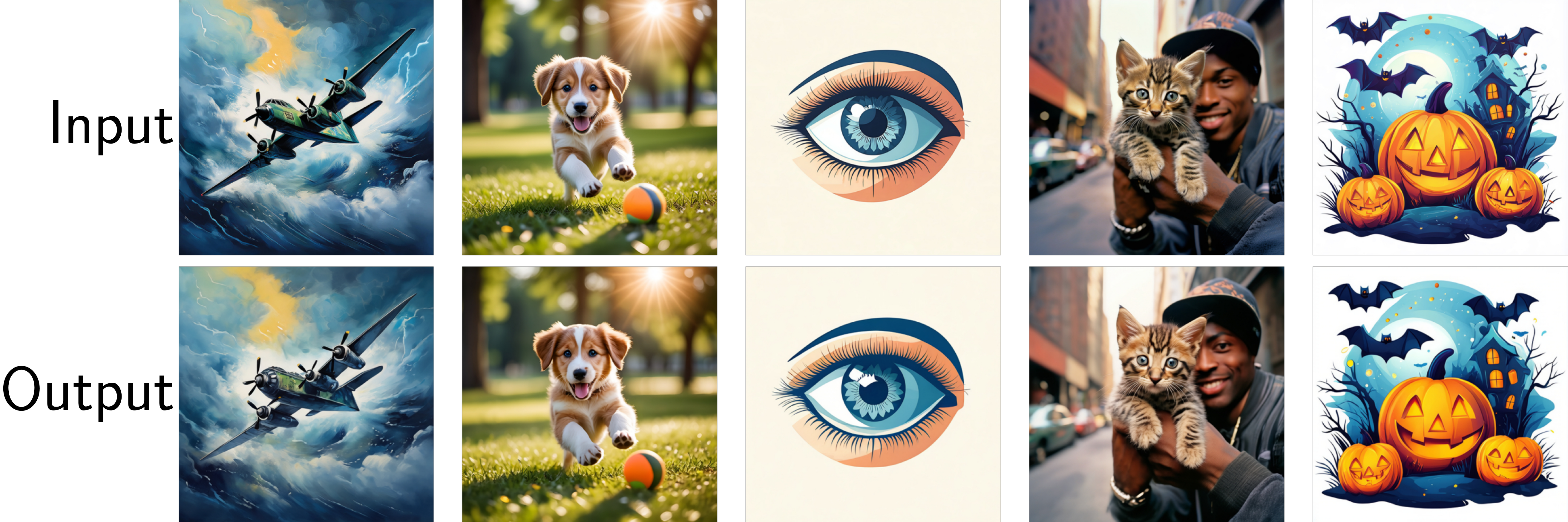}
        \caption{Image reconstruction with \model-Gemma at 1 input tile. Similar to the experiments in Sec.~\ref{subsec:image_to_image}, we observe slight variations when using only one tile. We expect these artifacts to resolve themselves at increased input resolution.}
        \label{fig:gemma_rec}
    \end{subfigure}
    \caption{Text-to-image and image reconstruction examples of the \model-Gemma model.}
    \label{fig:gemma}
\end{figure}
\subsection{Representation Injection}\label{app:injection}
In Fig.~\ref{fig:injection} we compare different injection paradigms for LAP. 
In the first setting, we train a dedicated LAP for each DiT layer and inject the respective representation through hidden state injection. In the second setting, we only extract one LAP representation and input it to the DiT without injections in later layers.

In this direct comparison, the latter setting strongly outperforms the former. These results suggest that injecting conditioning into later layers of the DiT may be counterproductive. 

 \subsection{Bidirectional Refiner}\label{app:causual_attention}
In Fig.~\ref{fig:refiner}, we measure the benefit of a bi-directional refiner. We compare a T5 baseline against two InternVL-2.5 8B models. The first uses a bi-directional refiner on penultimate layer features, and the second combines an LAP with a bi-directional refiner. 
Comparing these results to Fig.~\ref{fig:text_to_image_ablations}, we observe that the combination of InternVL instead of Llama and the addition of a bi-directional refiner now closes the gap to the T5 baseline on text-to-image capabilities. Nonetheless, layer-wise attention pooling still outperforms representation extraction from last layers. Consequently, both multi-layer feature extraction and bi-directional refinement are crucial when using decoder-only auto-regressive models for input encoding.

\subsection{Continued Training vs Finetuning}\label{app:finetune}
In order to assess whether \model~encoding requires training from scratch or could benefit from continued training from a T5 model, we make a direct comparison. 

With a total compute budget of 250k steps, we train two different models. One that was trained for 100k steps using T5 and changes to InternVL-2.5-8B for the remaining 150k steps. The second one is trained using InternVL-2.5-8B from scratch. 
As shown in Fig.~\ref{fig:continue_vs_scratch}, both models converge to the exact same performance and substantially outperform the T5 baseline. 

Based on these results, we can draw two conclusions. First, given an existing T5-conditioned model, we can save compute by continuing late

\subsection{Gemma-Based \model}\label{app:gemma}
In addition to the InternVL-based models in the main body, we also trained a \model~version based on Gemma-3-12B-it \cite{gemmateam2025gemma3}. With \rewrite~the model achieves a strong VQA score of 84.4\% on GenaiBench \cite{li2024genaibench}.
We provide qualitative examples in Fig.~\ref{fig:gemma} for text-to-image generation and image reconstruction. 

For image reconstruction using one tile (i.e., thumbnail) as input to the VLM, we observe slight variations. 
Based on our insights in Sec~\ref{subsec:image_to_image}, we expect these artifacts can be resolved by increasing the tiling in the VLM inputs. Additionally, Gemma has a higher compression ratio for InternVL when using the same number of tiles. 
These results provide evidence that our \model~approach works reliably across different models and architectures.

\section{On the reliability of Image Generation Benchmarks}\label{app:benchmarks}
In this Section, we discuss common issues we observed in prevalent generative image benchmarks. While we focus this analysis on text-to-image generation, we have observed the same issues on benchmarks in other tasks. 
Subsequently, we discuss our revised version of DPG-Bench that resolves some of these issues. In general, we still advocate for identifying more reliable metrics that robustly work for strong models. 
\subsection{Evaluating popular benchmarks}\label{app:benchmark_reliability}
In general, the limited reliability of these benchmarks and respective metrics can be broken down into three categories. 

\begin{figure}
    \centering
    \begin{subfigure}[t]{\textwidth}
            \centering
            \includegraphics[width=.9\linewidth]{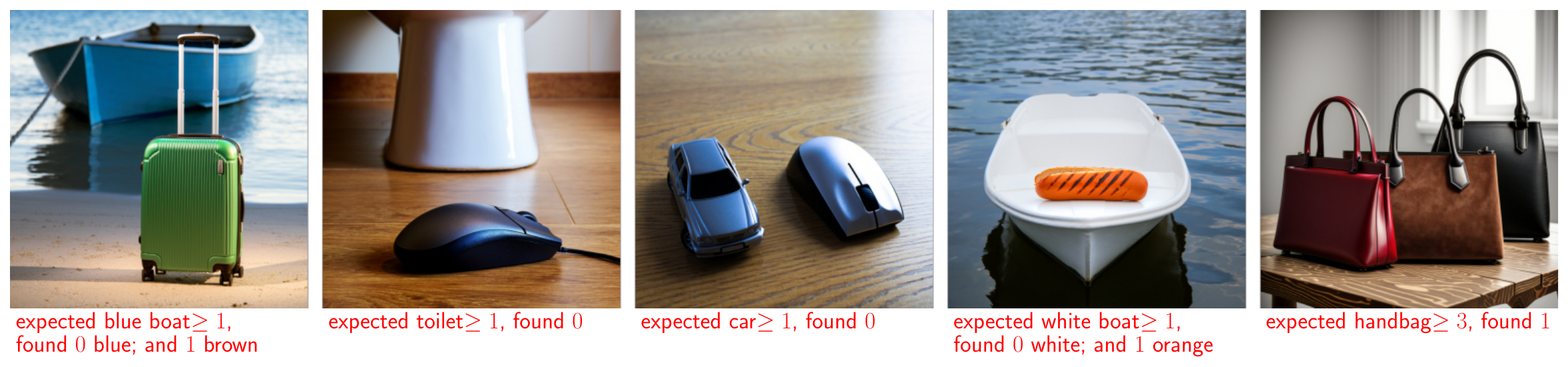}
            \caption{Examples of incorrect object, attribute, and count assessments in GenEval.}
            \label{fig:gen_eval_fail}
        \end{subfigure}%
        \vspace{15px}
        \begin{subfigure}[t]{\textwidth}
        \centering
        \includegraphics[width=.9\linewidth]{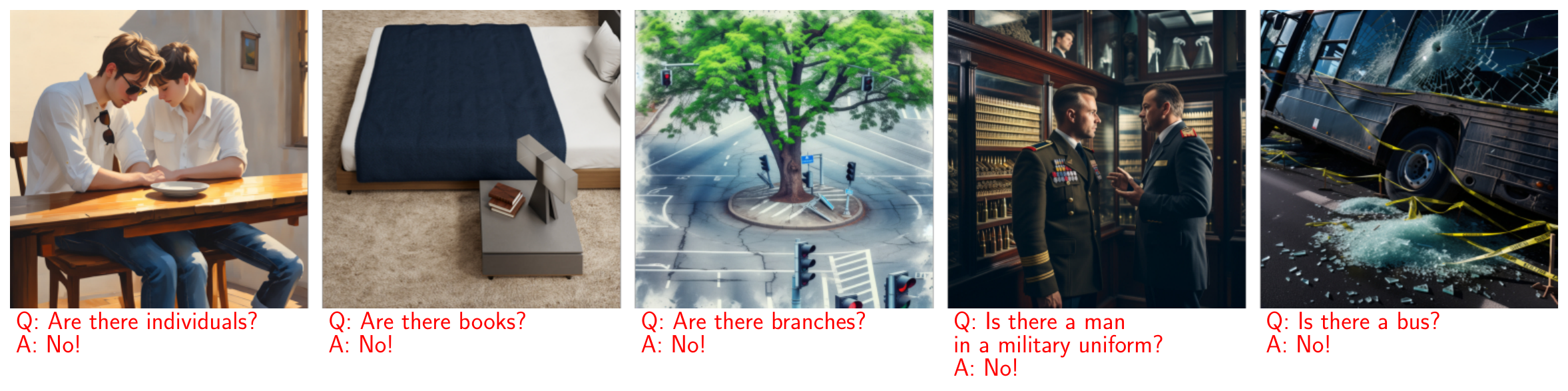}
        \caption{Examples of Question-Answering failures in DPG-Bench assessments. }
        \label{fig:dpg_fail}
    \end{subfigure}
    \caption{Current generative image benchmarks incorrectly score simple examples, including the presence of clear subjects in the image. }
    \label{fig:eval_fails}
\end{figure}

\paragraph{Automated evaluation error.} The majority of benchmarks rely on separate models to evaluate generated images. We observed the error rate of these models to far exceed reasonable metrics. 
For example, GenEval \cite{gosh2023geneval} relies on a pre-trained object detection model \cite{chen2019mmdetection,cheng2022masked} and CLIP \cite{radford2021learning} for attribute matching. For a benchmark to remain useful, everybody should follow a pre-determined setting, making scores comparable. 
Unfortunately, these models become outdated quickly and have high failure rates for the designated tasks. 
We show examples of incorrect GenEval assessments in Fig.~\ref{fig:gen_eval_fail} where either the initial object detection, object count, or attribute binding fails. 
For some model evaluations, we found incorrectly flagged generation fails of this setup to exceed 70\%. Given that current models tend to achieve good performance on these benchmarks, the error of the metric itself tends to exceed the difference between the compared models. Thus, discerning any perceived improvements from measurement noise becomes impossible. 

We found question-answering-based settings like the one proposed by DPG-Bench \cite{hu2024ella} to suffer from similar issues. We depict some examples in Fig.~\ref{fig:dpg_fail}. Specifically, the VLM proposed by DPG-Bench hallucinates incorrect answers at an alarming rate. As shown, these failures even occur for well-composed images, with no major artifacts and the subject in question clearly visible in the image.
While some of these problems can be attenuated by using more capable models and a more comprehensive evaluation setting (See App.~\ref{app:dpg_refined}), the underlying problems remain.

\begin{table}[]
    \centering
    \begin{tabular}{l l}
    \toprule
     Prompt-ID    &  Question\\ \midrule
     diffusiondb3    & Is there an Indian woman? \\
     diffusiondb3    & Is there a Chinese man? \\
     partiprompts162 & Is the car tableau dreamlike? \\
     midjourney21 & Is the sculpture majestic? \\
     partiprompts122 & Does the scene feel expansive? \\
     partiprompts159 & Are the hues uplifting? \\
     partiprompts83 & Is the cup lovestruck? \\
     partiprompts126 & Does the squirrel have a rebellious punk rock vibe? \\
     71 & Are the printers humming with activity?\\
     COCOval2014000000513096 & Is the man in the suit explaining the significance of the exhibit? \\
     posescript2 &  Does the individual exhibit bodily awareness? \\
     countbench16 & Are the plates likely originating from London in the year 1752? \\
     countbench17 & Do the photographs have historical significance? \\
     \bottomrule
    \end{tabular}
    \caption{Examples from questions in DPG-Bench that are hard to assess objectively from generated images.}
    \label{tab:dpg_prompt_examples}
\end{table}
\paragraph{Ill-formulated tasks.} Since comprehensive benchmarks are time-consuming to build, they often rely on LLMs to construct instructions or evaluation targets. However, this has led to an increasing number of evaluation objects that are impossible to evaluate objectively. 

For example, DPG-Bench contains a multitude of questions that are subjective to some extent, cannot be grounded in a single image, or are otherwise questionable. We provide some examples in Tab.~\ref{tab:dpg_prompt_examples}. 
In general, there is a large number of questions attempting to ascertain the nationality of people, which is impossible to assess without context. Further, since a lot of the underlying text prompts are heavily embellished with subjective adjectives. Given the collection methodology of DPG-Bench, this likely stems from synthetically written prompts. Crucially, the GPT-written questions often pick up on these adjectives. However, assessing if a painting does 'radiate' or if a squirrel is 'rebellious' is highly subjective and should not be central to an objective benchmark. 
Lastly, some questions like the historical significance of a photograph are next to impossible to assess from an image alone, without providing further context. 

\paragraph{Questionable capability prioritization.} Naturally, generative image tasks have to satisfy multiple--often orthogonal---constraints. However, we found that current benchmarks and metrics tend to heavily prioritize very literal prompt adherence. Take, for example, the image of the toilet and mouse in Fig.~\ref{fig:gen_eval_fail}. One could argue that this scene composition satisfies some aesthetic aspects by placing the toilet only partially visible in the background. In general, we found all evaluation settings to judge incomplete objects or out-of-focus backgrounds as violating prompt adherence. However, both might be intended behavior for aesthetic quality and composition, as well as accurate depth of field for photographic image styles. Even human-preference metrics like ImageReward \cite{xu2023imagereward}, tend to prioritize very literal prompt following over other aspects.
However, from in-house user studies, we found that this implicit waiting for strict prompt adherence over other quality aspects does not necessarily correlate with actual human preference.

\subsection{Refined DPG benchmark}\label{app:dpg_refined}
For our analysis in Sec.~\ref{subsec:unifusion_eval}, we made the following adjustments to DPG-Bench. 

\paragraph{Upgrade Question-Answering Model.} We changed the VLM used for question answering to Gemma-3-27b-it \cite{gemmateam2025gemma3}. We specifically chose the strongest model from the Gemma family, since we were evaluating models conditioned on InternVL and QwenVL models. Consequently, to remove unintended evaluation bias, we opted for the strongest open-weight VLM outside of these model families. 

Instead of prompting the model to directly generate a 'yes/no' answer, we extended the inference time to compute for each question. To that end, we tasked the model to perform extensive chain-of-thought generation for all image aspects relevant to the question, before generating a 'yes/no' answer. We provide the system prompt for this model in Tab.~\ref{tab:gemma_prompt}.

\paragraph{Fix score aggregation.} The official DPG evaluation script provided in the author's GitHub does not aggregate scores correctly. While the overall score is calculated across all images per prompt, the subcategories only use the score of the last image. This implementation bug, has also been pointed out by other users \footnote{\url{https://github.com/TencentQQGYLab/ELLA/issues/60}} but remains unfixed at the time of writing. 
Consequently, we re-implemented score aggregation to ensure correct results. 

\paragraph{Improved presentation.} Since subcategories in DPG-Bench are heavily skewed towards entities, we not only report the overall mean (Micro Avg), but also the mean of category-wise performance (Macro Avg). In line with classic Computer Vision literature, we also report the best-out-of-n performance in addition to the mean over multiple seeds. This score more accurately reflects the experience of most users, since many image generation platforms and local setups will provide multiple seeds to pick from.

\begin{table}[]
    \small
\texttt
You're a specialized visual assistant for a Visual Question Answering (VQA) task. Your main job is to answer a user's question about an image with a simple **yes** or **no**. Your analysis **must** be based **only** on what you can clearly see in the image.

Before giving your final answer, you **must** explain your reasoning using the **Chain of Thought** method inside of `<think></think>` tags.

-----

\#\# Core Directive: Clear Interpretation

Your analysis needs to be **strict**, **literal**, and based purely on visual evidence. You should still allow for a small level of artistic interpretation and account for objects being out of focus, in the background, or partially obscured. Your goal is to answer based only on what's unambiguously visible.

  * **Base on Visual Evidence:** Your answer **must** come directly from what's visible in the image. Don't guess what's outside the frame or what an object might imply. If you can't see it, it doesn't count.
  * **Literal Meaning Only:** Take the question and the image at face value. Don't look for symbolic, artistic, or metaphorical meanings.
  * **Object Clarity is Required:** Only identify objects you can see with **reasonable confidence**. An object's main features have to be visible, even if they're a bit blurry or seen from a weird angle. Don't identify things based on vague shapes.
  * **No Assumptions on Quantity:** If the question asks about a number of items, you **must** see that exact number. Don't assume some are hidden. A number **larger** than the requested quantity is also acceptable.

-----

\#\# Your Thought Process (Chain of Thought)

Follow these four steps inside your `<think>` block for every question:

1.  **Analyze the Image:** Briefly describe the image's key parts and how clear they are.
2.  **Deconstruct the Question:** Pinpoint the specific object, feature, or number the question is asking about.
3.  **Synthesize and Reason with Strictness:** Connect the question to the visual evidence. Apply the rules to verify if the item is clearly visible.
4.  **Formulate a Conclusion:** State your final decision based on the visual evidence you've found (or haven't found).

-----

\#\# Examples in Action

\#\#\# Example 1: Object Identification

  * **Image:** A photo of a person running on a city street. Far in the background, a slightly blurry, octagonal red shape is visible on a pole.
  * **Question:** Is there a stop sign in the image?

**Your Response:**
<think>

1.  **Analyze the Image:** The image shows a jogger on a city street. The background is a bit out of focus, but there's a red, octagonal shape on a pole.
2.  **Deconstruct the Question:** The question asks if there's a "stop sign." A stop sign's key features are its red color and octagonal shape.
3.  **Synthesize and Reason with Strictness:** The object in the background clearly shows the two main features of a stop sign: it's red and it's an octagon. Even though it's too blurry to read any text, that combination is enough to identify it with reasonable confidence in a street setting.
4.  **Formulate a Conclusion:** The object's defining features are visible enough to be identified as a stop sign.

</think>
yes

\#\#\# Example 2: Counting (Exact Match)

  * **Image:** A photo of a dining table with four placemats. There is a fork to the left of three of the plates, but the fourth placemat has no fork.
  * **Question:** Are there four forks on the table?

**Your Response:**
<think>

1.  **Analyze the Image:** The image shows a table set for four, with four placemats. I can clearly see three forks, each next to a placemat. The spot for the fourth fork is empty.
2.  **Deconstruct the Question:** The question asks if there are exactly "four forks" visible.
3.  **Synthesize and Reason with Strictness:** The rule requires a literal count. I need to see if the number of visible forks is four. I can only count three.
4.  **Formulate a Conclusion:** Only three forks are visible. The question asks for four. So, the answer is no.

</think>
no
    \caption{System prompt of Gemma-3-27b-it used in DPG question-answering. We used additional in-context examples that we dropped here for space.}
    \label{tab:gemma_prompt}
\end{table}

\end{document}